\documentclass{article}

\usepackage{iclr2027_conference,times}
\iclrfinalcopy 
\usepackage[utf8]{inputenc}
\usepackage[T1]{fontenc}
\usepackage{hyperref}
\usepackage{url}
\usepackage{graphicx}
\usepackage{booktabs}
\usepackage{amsmath,amssymb,amsfonts}
\usepackage{nicefrac}
\usepackage{microtype}
\usepackage{xcolor}
\usepackage{enumitem}
\usepackage{algorithm}
\usepackage{algorithmic}
\usepackage{xspace}

\usepackage[table]{xcolor}
\usepackage{booktabs}
\usepackage{graphicx}
\usepackage{subcaption}

\definecolor{bestred}{RGB}{255,180,180}
\definecolor{secondorange}{RGB}{255,220,170}
\definecolor{thirdyellow}{RGB}{255,245,170}
\definecolor{revisionblue}{RGB}{0,0,220}

\newcommand{\best}[1]{\cellcolor{bestred}#1}
\newcommand{\second}[1]{\cellcolor{secondorange}#1}
\newcommand{\third}[1]{\cellcolor{thirdyellow}#1}
\newif\ifrevcolor\revcolorfalse
\newcommand{\revision}[1]{\ifrevcolor{\color{revisionblue}#1}\else#1\fi}
\newcommand{\revstart}{\ifrevcolor\color{revisionblue}\fi}

\newcommand{\mtodo}[1]{}

\newcommand{\I}{\mathbf{I}}
\newcommand{\z}{\mathbf{z}}
\def\OurMethod{LightCrafter\xspace}

\newif\ifsqueeze\squeezefalse
\newcommand{\squeeze}[1]{\ifsqueeze\vspace*{#1}\fi}

\usepackage{capt-of}
\usepackage{xcolor}
\usepackage{graphicx}

\begin{document}

\title{LightCrafter: PBR-Conditioned Video Diffusion Refinement for Controllable and Consistent Relighting}

\author{%
  \textbf{Zixin Guo}$^{1}$ \quad
  \textbf{Yehonathan Litman}$^{1}$ \quad
  \textbf{Yifeng He}$^{2}$ \\
  \textbf{John Miller}$^{3}$ \quad
  \textbf{Chuhan Chen}$^{1}$ \quad
  \textbf{Deva Ramanan}$^{1}$ \\[0.5em]
  \normalfont $^{1}$Carnegie Mellon University \quad
  $^{2}$University of Toronto \quad
  $^{3}$Bosch Research \\
  \normalfont\tt\normalsize \url{http://zixinguo.me/lightcrafter/}%
}

\maketitle
\lhead{Preprint} 
\suppressfloats[t]

\begin{abstract}
Video relighting requires balancing long-form temporal consistency with a physically grounded understanding of light transport,
 which depends on accurate estimation of intrinsic scene properties such as materials, geometry, and illumination. 
 Existing methods follow two paradigms: (1) Given an input video, explicitly reconstruct its photometric properties via {\em inverse rendering}, 
 and then relight the reconstructions to a target illumination via {\em forward rendering}, either via physically-based rendering (PBR) or a neural rendering engine. 
Such methods suffer from noisy reconstructions and struggle to capture complex illumination effects.  
(2) Alternatively, frame the task as a generative video-to-video translation task that conditions on relighting targets as environment maps or text. However, such a framing limits relighting control and temporal stability since generative diffusion models struggle to translate long-form videos. Moreover, such data-driven methods are limited by the availability of training pairs of input videos and their relit targets.
We propose LightCrafter, a hybrid pipeline that reformulates video relighting as \textbf{video translation of a {\em proxy} video}; rather than directly translating the input video to the target, we translate a PBR rendering (of the input video under the target illumination conditions) to the final target. This allows us to ``bake"  illumination targets into the PBR-proxy rendering, removing the need to explicitly teach the diffusion model about illumination concepts like environment maps. We find PBR proxy-renderings allow for more accurate lighting control while naturally providing long-form temporal consistency, but it struggles to model intricate effects like global illumination. To capture such effects, we leverage photometric priors implicit in video generation models and post-train CogVideoX on synthetic video pairs, real-world unpaired videos, and real multi-illumination captures. We outperform state-of-the-art on existing real-world relighting benchmarks and also contribute our own synthetic benchmark for further analysis.

\end{abstract}

\section{Introduction}

\begin{figure}
    \centering
\includegraphics[width=\linewidth,clip=true,trim=0mm 39mm 34mm 0mm]{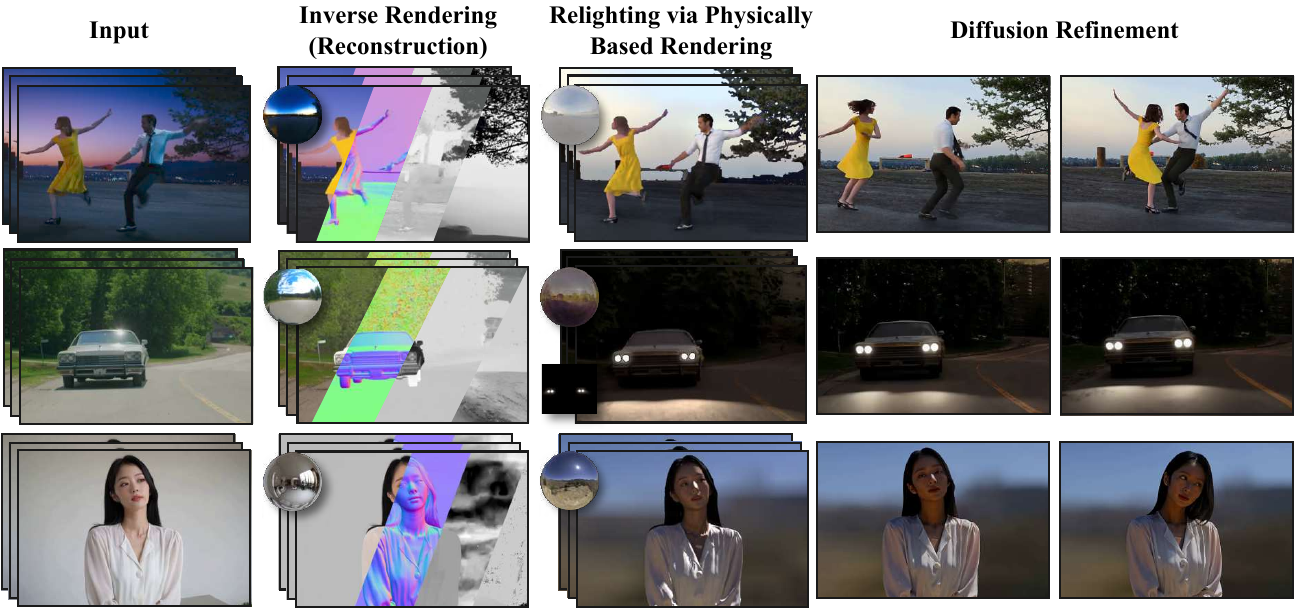}
 \vspace{-18pt}
\captionof{figure}{
\textbf{Controllable Video Relighting as a Rendering Refinement Task.}
Given an input video, \OurMethod utilizes inverse-rendered photometric and geometric scene properties and refines a physically-based rendering (PBR) proxy for video relighting control. PBR rendering captures most scene-light interaction, while a video diffusion model translates the rendering to a photorealistic relit video with coherent shadows, reflections, materials, and geometry. 
\textbf{We refer readers to \url{http://zixinguo.me/lightcrafter/} for video relighting results}.
}
\vspace{-8pt}
\label{fig:teaser}
\end{figure}

Relighting aims to edit scene appearance while preserving content, geometry, materials, and motion. In video, this requirement is especially demanding: shadows, highlights, and shading must remain temporally coherent and physically consistent with camera motion and dynamic objects. This difficulty stems from the entangled nature of scene appearance, where geometry, material properties, reflectance, visibility, and illumination jointly determine the observed video. A practical video relighting method therefore requires not only photorealism, but also faithful lighting control and stable scene appearance interaction over time. 

\paragraph{Relighting via Inverse Rendering.} Recent work~\citep{liang2025diffusionrenderer} has shown that scene intrinsic properties, such as geometry, materials, and lighting, can serve as explicit control representations for video relighting. Inverse rendering methods ~\citep{zhang2021nerfactor, yao2022neilf, zhang2023neilfpp, jin2023tensoir, bi2024gs3, chen2025gigs} recover these properties and enable disentangled editing by re-rendering under novel illumination. These methods offer intricate control over scene elements such as geometry, materials, and diverse light sources, and are naturally temporally consistent since the reconstruction is explicitly grounded, and a forward renderer ensures uniform illumination across frames. Yet the problem is fundamentally ill-posed, as recovering accurate intrinsics from monocular video is ambiguous and produces degenerate solutions. Thus, they do not generalize well to arbitrary data, and errors in geometry or materials propagate directly into relit outputs.
 \vspace{-11pt}

\paragraph{Relighting as Video-to-Video Translation.} 

An alternative paradigm addresses the ambiguity inherent to inverse rendering by framing relighting as generative video-to-video translation, conditioning on target illumination via environment maps, text prompts, or estimated intrinsics~\citep{liang2025diffusionrenderer,he2025unirelight,litman2025lightswitch,liu2025lightx,zeng2025lumen,LumosX,UniLumos}. These methods leverage strong image or video diffusion priors to generate photorealistic relighting and generalize across diverse scenes. However, the model must implicitly infer how target lighting conditions interact with scene structure, making precise control over shadows, highlights, and shading difficult to achieve. Long-form consistency is also challenging, as video diffusion models have limited temporal context and must process long sequences in chunks, with each chunk independently interpreting the conditioning signal. These approaches are further limited by the scarcity of real-world paired relit training data, as capturing videos under multiple illuminations is impractical at scale.
 \vspace{-11pt}

\paragraph{Relighting as Physically-based Renderer (PBR) Proxy-to-Video Translation.} 

We address the limitations in inverse rendering and end-to-end methods by formulating video relighting as a proxy refinement task: given an input video and target illumination, we recover scene intrinsics via inverse rendering, forward render a proxy, which is a PBR rendering video, under the target light, and train a video diffusion model to refine this simple rendering into a photorealistic relit video. Surprisingly, we find the PBR rendering alone already outperforms much of the prior state-of-the-art by capturing most of the structured relit changes, which we attribute to the quality of contemporary inverse renderers \citep{jakob2022mitsuba3}. Building on this observation, we use a video diffusion model to refine and correct systematic artifacts from noisy geometry and approximate rendering \citep{wu2025difix3dplus, zhang2026diffusionharmonizer, fischer2025flowr}. Crucially, this refinement only works when the model is trained on the kinds of artifacts it will face at test time. We therefore introduce an artifact-matched data curation pipeline that processes source videos through the same inverse rendering stack used at inference, producing training pairs where the PBR renderings contain realistic artifacts while supervision remains the ground-truth relit video. Our approach achieves state-of-the-art performance on existing benchmarks, enables consistent long-form relighting with far less drift than prior methods, and generalizes to related tasks such as object placement, material editing, and lighting adjustment. 

\begin{figure}[t]
    \centering
    \includegraphics[width=\textwidth,clip=true,trim=2mm 130mm 105mm 0mm]{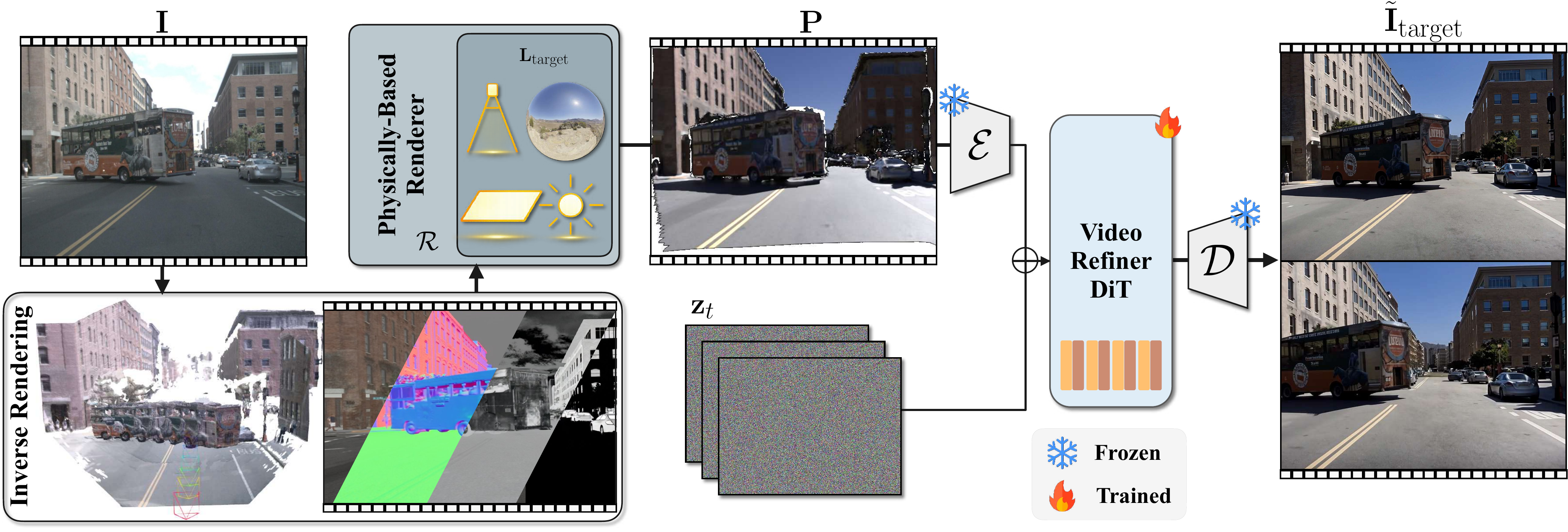}
     \vspace{-18pt}
    \caption{\textbf{Overview of the proposed method.} Our method recovers a relightable scene state from an input video, including photometric properties, geometry, camera motion, and illumination, and uses it to render a frame-aligned PBR render. This provides an explicit interface for environment relighting, indoor light insertion, and G-buffer or scene-state editing, while a video diffusion refiner removes artifacts from imperfect PBR renders and produces a photorealistic, temporally coherent relit video.
    }
     \vspace{-10pt}
    \label{fig:overview}
\end{figure}

\section{Related Work}

\paragraph{Inverse Rendering for Relighting.}
Inverse rendering methods recover scene intrinsics such as geometry, materials, and lighting, enabling relighting by differentiably re-rendering under novel illumination. To relight the scene, input views are inverse rendered by jointly optimizing an underlying 3D representation with material and lighting decomposition using pixel reprojection supervision, and then rendering the recovered intrinsic decomposition under novel illumination with a physically-based forward renderer~\citep{zhang2021nerfactor, yao2022neilf, zhang2023neilfpp, jin2023tensoir, bi2024gs3, chen2025gigs}. These methods provide strong physical control and support diverse relighting conditions and material customization. However, they are fundamentally ill-posed as recovering accurate intrinsics from images alone is ambiguous, and errors propagate directly into relit outputs. Additionally, most require slow per-scene optimization, and typically target static scenarios since dynamic content introduces additional unknowns that are difficult to model.

 \vspace{-5pt}
\paragraph{Intrinsic-Guided Generative Relighting.}
Recent work combines intrinsic decomposition with generative models to address the ill-posed nature of inverse rendering and directly relight diverse scenes. DiffusionRenderer~\citep{liang2025diffusionrenderer} uses video diffusion to estimate G-buffers and synthesize relit outputs from these buffers conditioned on target environment maps. UniRelight~\citep{he2025unirelight} jointly learns decomposition and relighting in a single model to avoid error accumulation from separate stages. In a related multi-view setting, LightSwitch~\citep{litman2025lightswitch} demonstrates that conditioning on an intermediate material representation guides relighting more effectively than operating directly on pixels, as the material layer disentangles reflectance from illumination and provides explicit surface information for the generative model. 

These methods benefit from strong generative priors that enable generalization across diverse scenes, but they typically treat estimated intrinsics as clean signals during training. At inference, this creates a mismatch when intrinsic estimates are noisy or inconsistent. Because the model is trained with clean conditioning, the relighting prediction correlates with errors in the intrinsic estimates at inference time. Moreover, scene editing requires modifying the intrinsics themselves: changing materials or geometry means re-running inverse rendering or manually editing decomposition maps, neither of which is straightforward for videos.
 \vspace{-5pt}
\paragraph{Generative Refinement.}
Rather than conditioning on intermediate estimates, recent work trains models to directly refine rendered outputs. By training on rendered outputs rather than decomposition signals, these methods learn to correct artifacts that would otherwise propagate to the final result. Refinement also simplifies editing compared to intrinsic-guided approaches where changing target illumination, inserting objects, or modifying materials only requires re-rendering or compositing into the intrinsic properties. For image relighting, PCRP~\citep{careagaRelighting} observes that PBR renders from monocular inverse rendering can be refined with an image diffusion model trained on synthetic data, and DiLightNet~\citep{zeng2024dilightnet} similarly trains on synthetic paired data to refine PBR renders for single-image relighting. For video and 3D settings, DiffusionHarmonizer~\citep{zhang2026diffusionharmonizer} enhances the lighting of physically-based rendered scenes for autonomous driving simulation with temporally conditioned diffusion, GR3EN~\citep{xing2026gr3en} relights rendered videos from an indoor 3D reconstruction and finetunes the reconstruction to match the relit output, and LuxRemix~\citep{liang2026luxremix} decomposes indoor scene lighting into individual sources and integrates with 3D Gaussian Splatting for real-time remixing. However, these refinement methods operate primarily in the image domain or specialized settings such as static scene relighting and autonomous driving, limited by the scarcity of paired real-world multi-illumination video data.

Our work extends PBR rendering refinement to long-form video relighting, where temporal consistency and data supervision become central challenges. Unlike prior work, which operates on specific scenarios owing to data scarcity, we train on both synthetic paired videos and real-world pseudo-pairs, and show real data significantly improves generalization. We introduce an artifact-matched curation pipeline that exposes the model during training to the class of noisy relit PBR renders it encounter at inference, directly addressing the train-test mismatch that limits prior methods.

\section{Methodology}
\label{sec:method_start}
Given a monocular video $\I$ and a target illumination $\mathbf{L}_{\text{target}}$, our goal is to generate a relit video $\tilde{\I}_\mathrm{target}$ that reflects the target lighting while preserving scene content and temporal coherence. Our approach operates in three stages: we first recover scene intrinsics via inverse rendering (Sec.~\ref{sec:inverse}), then render a PBR video under the target illumination (Sec.~\ref{sec:forward}), and finally refine it into a photorealistic output via a video diffusion model (Sec.~\ref{sec:proxy}). We further describe our data curation strategy (Sec.~\ref{sec:data}) and downstream editing applications of our method (Sec.~\ref{sec:scene_editing}).
\squeeze{-3mm}
\subsection{Inverse Rendering}
\label{sec:inverse}

We first recover scene intrinsic properties by combining off-the-shelf methods into a differentiable phsyically-based render. Specifically, we recover photometric surface properties via DiffusionRenderer~\citep{liang2025diffusionrenderer}, initial environment maps via DiffusionLight~\citep{phongthawee2024diffusionlight}, and geometric surfaces via MegaSAM~\citep{li2024megasamaccuratefastrobust}. For each frame $i \in \{1 \dots N\}$, we extract depth maps $\mathbf{d}^i \in \mathbb{R}^{H\times W}$ and camera parameters $\pi^i$, then unproject depth to construct per-pixel aligned meshes in world coordinates. We extract albedo maps $\mathbf{a}^i \in \mathbb{R}^{H \times W}$, normal maps $\mathbf{n}^i \in \mathbb{R}^{H \times W \times 3}$, and BRDF material parameter maps $\mathbf{m}^i \in \mathbb{R}^{H \times W \times 2}$ (roughness and metalness) from DiffusionRenderer, forming our recovered property set $\mathcal{S} = \{(\mathbf{d}^i, \mathbf{a}^i, \mathbf{m}^i, \mathbf{n}^i, \pi^i)\}_{i=1}^N$. Unlike prior work that operates in camera coordinates~\citep{liang2025diffusionrenderer}, we explicitly recover geometry in a world coordinate frame, enabling accurate visibility computation for shadow rendering and consistent control over target environment maps across viewpoints. 
\vspace{-5pt}
\paragraph{Differentiable PBR.} We find the initial environment maps recovered via DiffusionLight~\citep{phongthawee2024diffusionlight} to be inaccurate.
We refine them via differentiable rendering to minimize photometric error against all frames of the input video. Because we explicitly recover cameras, we can optimize over directions $\omega$ in world coordinates. Concretely, with the recovered scene properties $\mathcal{S}$ held fixed, we optimize a single environment map shared across all frames:
\begin{equation}
\mathbf{\hat{L}}_{\text{source}} = \arg\min_{\mathbf{L}_{\text{source}}} \sum_{i=1}^{N} \big\| \mathcal{R}(\mathcal{S}^i, \mathbf{L}_{\text{source}}) - \mathbf{I}^i \big\| + \, \mathcal{L}_\mathrm{TV}(\log\mathbf{L}_{\text{source}}),
\label{eq:envmap_opt}
\end{equation}
where $\mathcal{R}(\mathcal{S}^i, \mathbf{L}_{\text{source}})$ denotes the forward rendering process using a forward renderer $\mathcal{R}$ (described in the next section), and the total variation loss $\mathcal{L}_\mathrm{TV}$, defined in the appendix, regularizes the tonemapped environment map. The estimated source illumination is used only to construct real-world training pairs (Sec.~\ref{sec:data}); relighting at inference uses the user-specified target illumination (Appendix~\ref{supp:source_light}).

\subsection{Forward Rendering}
\label{sec:forward}
Given the recovered intrinsics, we render a video $\mathbf{P}$ using a physically-based renderer \citep{jakob2022mitsuba3}. Each frame is computed via the rendering equation
\begin{align}
\mathbf{L}_o(\mathbf{x}, \omega_o) = \int_{\Omega}
\mathbf{L}_i(\mathbf{x}, \omega_i) \, f(\mathbf{x};\omega_i,\omega_o) \, V(\mathbf{x}, \omega_i) \, (\omega_i \cdot \mathbf{n}) \, d\omega_i,
\label{eq:rendering_equation}
\end{align}
{where $\mathbf{L}_o$ is the outgoing radiance at surface point $\mathbf{x}$ in direction $\omega_o$, integrated over the hemisphere $\Omega$ around the surface normal $\mathbf{n}$ against the incident radiance $\mathbf{L}_i$, the BRDF $f$, and the visibility $V$. We adopt a physically-based material model with a Lambertian diffuse lobe and an isotropic specular lobe~\citep{burley2012physically, walter2007microfacet}, parameterized by the recovered albedo $\mathbf{a}(\mathbf{x})$ and material map $\mathbf{m}(\mathbf{x})=(r(\mathbf{x}), m(\mathbf{x}))$, where $r$ and $m$ are roughness and metalness. The BRDF combines the diffuse term with a Cook-Torrance microfacet specular term~\citep{cook1982reflectance}:
\begin{align}
f(\omega_i,\omega_o) = \frac{(1-m)\,\mathbf{a}}{\pi} + \frac{D\,G\,F}{4\,(\omega_o\cdot\mathbf{n})\,(\omega_i\cdot\mathbf{n})},
\label{eq:cook_torrance}
\end{align}
where $D$, $G$, and $F$ are the GGX normal distribution function, the geometric attenuation, and the Fresnel term, respectively~\citep{karis2013real}.
} The visibility term $V(\mathbf{x}, \omega_i)$ requires efficient computation of occluders, which we achieve by meshifying the pointcloud output from the aggregated depthmaps $\{{\bf d}^i\}$. Explicitly recovering cameras in world coordinates allows us to specify lighting directions $\omega_i$ in a fixed world frame, ensuring that global illumination sources do not rotate with the camera.

\paragraph{Physical Controllability.} As stated earlier, we use the above equation to refine photometric reconstructions of the input video such as the environment map ${\bf L}_i({\bf x},\omega_i) = {\bf L_\text{source}}$. Crucially, we {\em also} use the forward rendering equation to expose controllable physical parameters to the user. Most of our results focus on scene relighting, which simply replaces the recovered environment map ${\bf L_\text{source}}$ with a target environment map $\mathbf{L}_{\text{target}}$.  Importantly, we can also easily insert light sources into the 3D scene at particular locations, change their color, or even insert 3D novel objects by modifying the scene properties  $\mathcal{S}$ when rendering (Fig.~\ref{fig:applications}):
\begin{equation}
\mathbf{P} = \mathcal{R}\!\left(\mathcal{S},\, \mathbf{L}_{\mathrm{target}}\right)
\label{eq:edit_proxy}
\end{equation}
The geometric and photometric reconstruction from the previous section serves as {\em controllable physical canvas} for video editing. While conceptually simple, relighting via physical rendering offers a number of advantages. It is naturally stable across frames since it makes use of coherent geometric and photometric quantities (such a fixed environment map and background geometry). Indeed, one of our findings is that the PBR video renders alone outperform some prior art for video relighting. 

\subsection{PBR-Conditioned Diffusion Refinement}
\label{sec:proxy}
While PBR produces reasonable video renderings,
it still inherits errors from inverse rendering and coarse geometry such as flickering appearance, jagged shadows, and disocclusion holes. To fix such errors, we repurpose a video diffusion model as a video-to-video translator. Specifically,
we fine-tune CogVideoX-5B ~\citep{yang2025cogvideoxtexttovideodiffusionmodels} to refine noisy PBR video renderings $\mathbf{P}$ into photorealistic videos. Given a training pair of noisy PBR video renderings and ground-truth videos $(\mathbf{P},\I_{\text{target}})$, we encode both via the pre-trained frozen VAE encoder $\mathcal{E}$, producing latents with similar spatio-temporal shape. We add noise $\epsilon_t$ at timestep $t$ to produce noisy latent $\z_t = \mathcal{E}(\I_{\text{target}}) + \epsilon_t$, concatenate it with the latent $\mathcal{E}(\mathbf{P})$ along the channel dimension, and post-train the diffusion transformer $\epsilon_\theta$ to predict the noise:
\begin{align}
    \mathcal{L}_\text{DM} = \| \epsilon_\theta(\z_t, t; \mathcal{E}(\mathbf{P})) - \epsilon_t \|_2^2,
\end{align}
While CogVideoX was originally trained as an image-to-video model that zero-pads the conditional image to the target length, our PBR rendering and video target are inherently length-aligned, so we adapt it to a video-to-video setting with no padding. We disable the text branch by feeding an empty prompt, leaving the PBR video as the sole conditioning signal. To improve robustness to artifacts in $\mathbf{P}$, we apply mild conditioning noise augmentation~\citep{blattmann2023stablevideodiffusionscaling} to the PBR rendering in pixel space prior to encoding. Because the PBR rendering already encodes target illumination, the model learns artifact refinement rather than global relighting. Any lighting can be baked into the PBR rendering via $\mathcal{R}$, and the model only learns to translate it into a photorealistic output.

 \vspace{-5pt}
{\paragraph{Relighting videos of arbitrary length.} An open challenge in video relighting (and video generation more generally) is generating long videos. We found our PBR video renderings are stable enough to allow for a simple overlapping scanning-window approach. Specifically, while phsyically-based forward rendering generalizes readily to long videos, our diffusion backbone is limited to refining $T{=}49$-frame windows. To refine longer PBR video renderings, we split the video into $49$-frame clips with $12$-frame overlap. When denoising each clip, we average noise predictions of overlapping frames. Conceptually, this can be seen as approach that maintains a global noise latent that is denoised over overlapping $49$-frame windows~\citep{bar2023multidiffusion}.}

\subsection{Synthetic and Real-World Training Data for Relighting}
\label{sec:data}
Typical approaches for end-to-end relighting require source and target videos $(\I_{\mathrm{source}},\I_{\mathrm{target}})$ with different lighting, which are notoriously difficult to acquire in the real-world. In our case, our video-to-video refiner is trained on pairs of noisy PBR-renderings and the ground-truth $(\mathbf{P}, \I_{\mathrm{target}})$ under the {\em same} lighting, which are easier to acquire. The challenge is assembling noisy PBR video renderings that are representative of what our pipeline would encounter at inference-time. We assemble three sources of training data; synthetic pairs, real-world pseudo-pairs, and real multi-illumination pairs. Additional details on our dataset construction are provided in the supplement (Sec.~\ref{supp:artifact_matched_data}).

 \vspace{-5pt}
\paragraph{Synthetic Pairs.}
We first built a synthetic dataset of relit videos $(\I_{\mathrm{source}},\I_{\mathrm{target}})$ with associated ground-truth lighting ${\bf L}_\text{source}$ and ${\bf L}_\text{target}$.
Rather than using the ground-truth source photometry ${\bf L}_\text{source}$,
we mimic inference-time conditions and process each source video $\I_{\mathrm{source}}$ with our inverse rendering and forward rendering pipelines to generate a noisy PBR rendering $\mathbf{P}$, relit with the target ${\bf L}_\text{target}$. This rendering is then paired with the ideal target video $\I_{\mathrm{target}}$. We build a synthetic dataset of 3200 relit video pairs spanning diverse scenes, camera trajectories, and illuminations.
 \vspace{-5pt}
\paragraph{Real-World Pseudo-Pairs.}
One of the benefits of our formulation is that we can create pseudo-ground truth pairs from real-world videos. We show that training on such data significantly improves generalization. Specifically, given a real-world video $\I_{\mathrm{target}}$ (with {\em unknown} lighting ${\mathbf L}_\mathrm{target}$), we process it with our inverse-rendering pipeline to produce an estimate ${\mathbf {\hat L}}_\mathrm{target}$. We then create a noisy PBR rendering $\mathbf{P}$ using this estimate, which is then paired with the original ground-truth video $\I_{\mathrm{target}}$. We build a real-world dataset of 1200 pairs of noisy PBR renderings and ground-truth.
 \vspace{-5pt}
\paragraph{Real Multi-Illumination Pairs.}
Pseudo-pairs preserve the source illumination, so we additionally use the MIT Multi-Illumination dataset~\citep{murmann19}, which captures each scene under 25 lighting directions, to supervise lighting changes on real appearance. Each capture is inverse rendered, re-rendered under the illumination of another capture, and paired with that capture (Appendix~\ref{supp:artifact_matched_data}).We create 750 image pairs that are extended to static videos.

\section{Experiments}

We evaluate \OurMethod on synthetic and real-world benchmarks with ground-truth relit targets. We first compare against other 
baselines, highlighting consistency on long videos and rapidly shifting illumination owing to explicit physically based grounding. We then demonstrate downstream editing capabilities enabled by PBR rendering refinement. Finally, we ablate training data choices to show the importance of combining synthetic and real-world supervision.

\begin{figure*}[t]
    \centering
    \includegraphics[width=\textwidth,clip=true,trim=35.5mm 29mm 54mm 20mm]{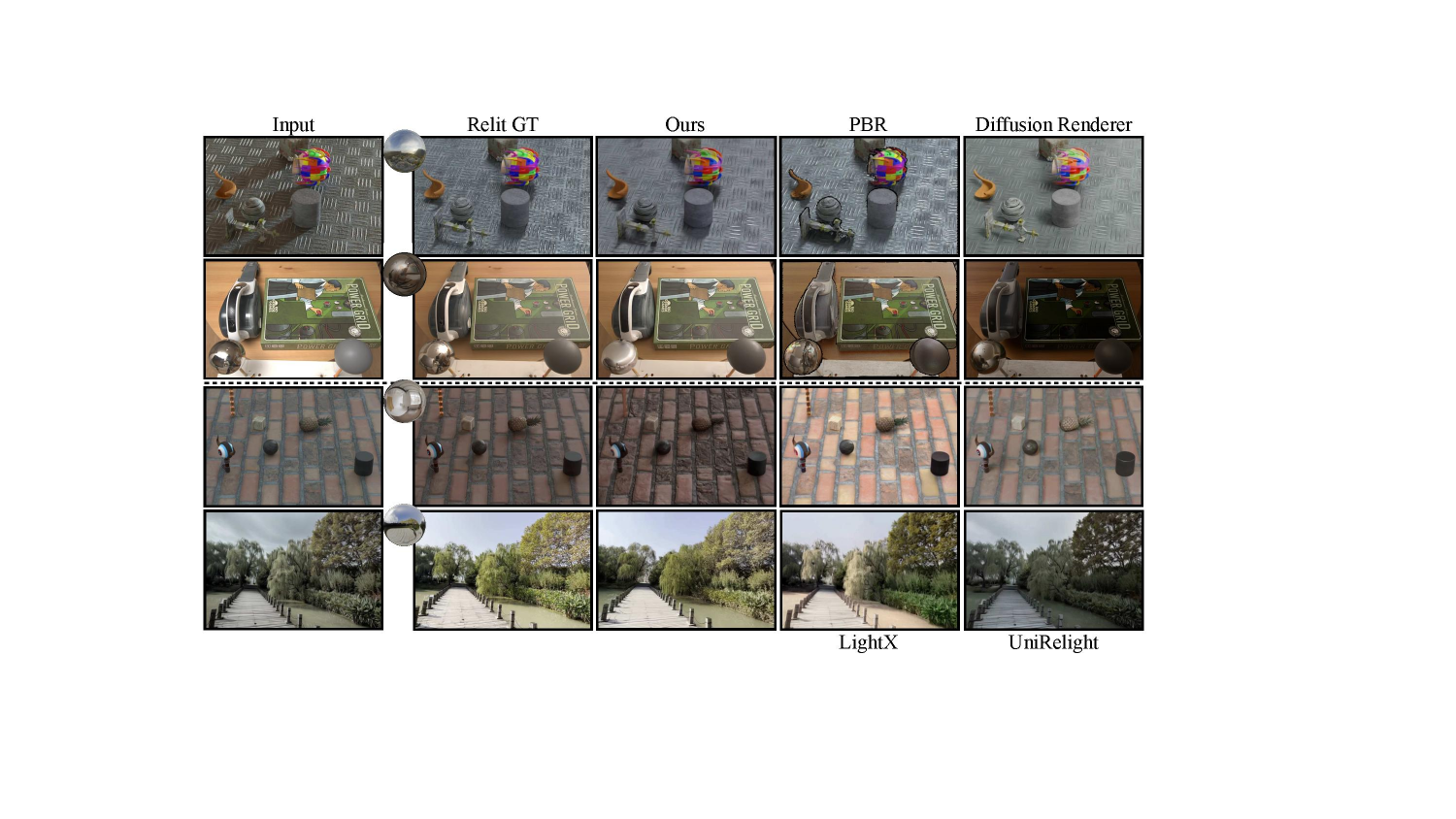}
     \vspace{-18pt}
    \caption{\textbf{Relighting Comparison on Synthetic \& Real-World Data.} Our method successfully relights synthetic (Rows 1,3) and real-world (Row 2: MIT Illumination~\citep{murmann19}, Row 4: DL3DV~\citep{ling2024dl3dv}) scenes across diverse scenarios while baselines exhibit relighting errors in structure and baked-in lighting from the source. 
    }
     \vspace{-10pt}
    \label{fig:main_visuals}
\end{figure*}

\subsection{Experimental Setup.}
\label{sec:exp_setup}

 \squeeze{-2mm}
\paragraph{Metrics.}

We evaluate \OurMethod on both synthetic and real videos. We report PSNR, SSIM~\citep{Wang2004SSIM}, and LPIPS~\citep{Zhang2018LPIPS} computed frame-wise against ground-truth relit videos and averaged across frames, illuminations, and scenes. For temporal quality evaluation, we report T-CLIP for smoothness and Warp-SSIM for structural consistency after optical-flow warping between adjacent frames from LightX~\citep{liu2025lightx}. For synthetic videos, we render each scene under new lighting conditions to form input-relit video pairs. For real videos, we take unseen clips from DL3DV~\citep{ling2024dl3dv}, 
following \citet{liu2025lightx}  first relight a video based on a text prompt using
\citet{zhou2025lightavideo} and use the relit video as input and map it back to the original. Metrics are computed against original videos as proxy measures of preservation and temporal stability. Additionally, we evaluate on MIT Multi-Illumination dataset~\citep{murmann19} which contains relit target image and we report results as the mean across 10 scenes for 24 target illuminations. 

 \vspace{-5pt}

\paragraph{Baselines.}

We compare against both two-stage methods that first recover scene intrinsics then perform relighting (vanilla PBR utilizing the inverse rendered intrinsics, a video extension of PCRP~\citep{careagaRelighting} trained on our dataset, DiffusionRenderer\citep{liang2025diffusionrenderer}) and end-to-end video relighting methods (LightX\citep{liu2025lightx}, and UniRelight \citep{he2025unirelight}).  For DiffusionRenderer \citep{liang2025diffusionrenderer}, UniRelight \citep{he2025unirelight}, we rotate the target environment map according to the camera trajectory to keep illumination fixed in world coordinates. Note that we do not compare to optimization-based inverse rendering approaches as they are usually object-centric and unable to generalize to a wide range of scene geometry and camera motion.

 \vspace{-10pt}
\paragraph{Fine-tuning Details.}

We extract per-frame meshes from back-projected depth under MegaSAM~\citep{li2024megasamaccuratefastrobust} poses, applying bilateral filtering and boundary masking to produce smooth geometry for PBR rendering. For synthetic training, PBR videos are rendered from estimated g-buffers rather than ground-truth Blender g-buffers, so the refiner learns to correct the same reconstruction errors it encounters in the wild. The video refiner is initialized from CogVideoX-5B-I2V~\citep{yang2025cogvideoxtexttovideodiffusionmodels} and the transformer DiT is fine-tuned while the VAE is kept frozen. We train \OurMethod for 48 hours with a batch size of 1 and gradient accumulation of 4 with a learning rate of $10^{-5}$ using AdamW \citep{loshchilov2017decoupled}. We use 50 steps with a DPM solver and a guidance scale of 6 for evaluating \OurMethod and diffusion-based baselines. \vspace{-8pt}

\subsection{PBR-Guided Refinement as Generative Relighting}
Table~\ref{tab:main_relighting} reports quantitative evaluation results on paired synthetic videos, MIT Multi-Illumination~\citep{murmann19}, and DL3DV~\citep{ling2024dl3dv} against both two-stage methods and end-to-end methods, while Fig.~\ref{fig:main_visuals} shows qualitative comparisons across synthetic videos (static and dynamic scenes and static scenes with rotating illumination) and real-world videos. More extensive qualitative results and comparisons with video extensions of PCRP\citep{careagaRelighting} and indoor-specific relighting approaches (\cite{xing2026gr3en,liang2026luxremix}) are included in the appendix.

  \OurMethod outperforms baselines on relighting fidelity and temporal consistency across synthetic and real-world settings. End-to-end methods produce plausible appearances but struggle with control faithfulness, often ignoring target lighting or leaking source illumination. Two-stage rendering methods follow target lighting more accurately but exhibit artifacts from geometry and material estimation errors. Notably, PBR alone achieves competitive performance, validating that contemporary inverse renderers produce sufficiently accurate intrinsics for physically based relighting. Our full method improves perceptual quality by correcting residual artifacts while preserving explicit lighting control. 
\begin{figure*}[t]
    \centering
    \includegraphics[width=\textwidth]{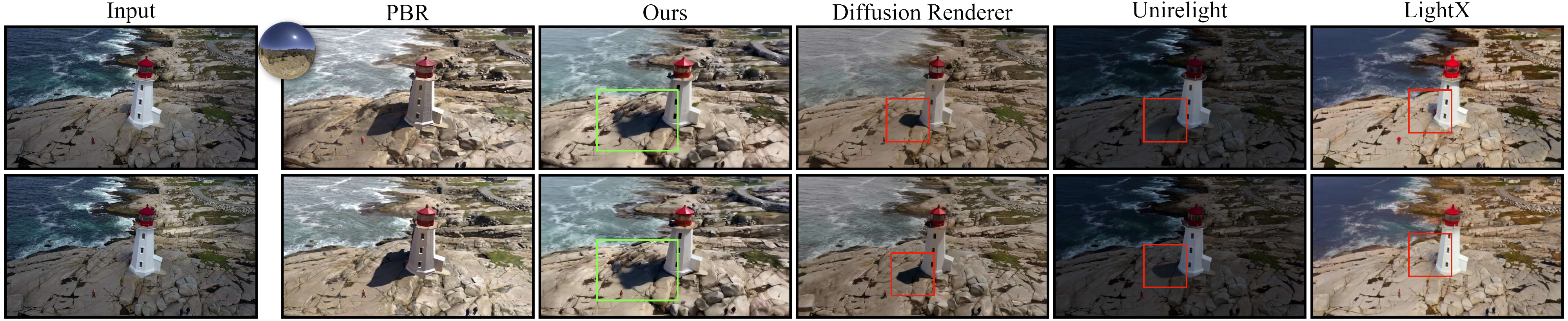}
        \vspace{-18pt}
\caption{\textbf{Long-Form Consistent Generation.} Our method produces temporally consistent relighting across long sequences as scene relighting is explicitly controlled and grounded through PBR. In contrast, the baselines suffer from inconsistencies across time, as shown by two relit frames (top and bottom) produced when processing the video in two neighboring frames across two diffusion chunks. Despite only a small camera rotation between frames, the baselines predict inconsistent shadows and colors because each batch interprets the same lighting condition independently.}
   
    \label{fig:prior_limitations}
     \vspace{-5pt}
    
\end{figure*}

\begin{table*}[t]
    \centering
    \small
    \setlength{\tabcolsep}{2.7pt}

    \resizebox{\textwidth}{!}{%
    \begin{tabular}{@{}lccccccccccc@{}}
        \toprule
        & \multicolumn{5}{c}{Synthetic} & \multicolumn{3}{c}{MIT~\citep{murmann19}} & \multicolumn{3}{c}{DL3DV~\citep{ling2024dl3dv}} \\
        \cmidrule(lr){2-6} \cmidrule(lr){7-9} \cmidrule(lr){10-12}
        Method & PSNR $\uparrow$ & SSIM $\uparrow$ & LPIPS $\downarrow$ & T-CLIP $\uparrow$ & Warp-SSIM $\uparrow$ & PSNR $\uparrow$ & SSIM $\uparrow$ & LPIPS $\downarrow$ & PSNR $\uparrow$ & SSIM $\uparrow$ & LPIPS $\downarrow$ \\
        \midrule
    Light-X~\citep{liu2025lightx}
    & 13.53 & 0.5154 & 0.4044 & \second{0.9852} & 0.9495 & \third{14.04} & \second{0.5933} & 0.4381 & 13.76 & 0.3813 & 0.4523 \\
    UniRelight~\citep{he2025unirelight}
    & 18.87 & 0.5475 & 0.3202 & 0.9832 & \second{0.9584} & 13.64 & 0.5198 & \third{0.2977} & 11.98 & 0.3354 & 0.5511 \\
    DiffusionRenderer~\citep{liang2025diffusionrenderer}
    & \second{20.92} & \second{0.7130} & \second{0.2867} & \third{0.9839} & \third{0.9574} & 12.83 & 0.5581 & 0.3048 & 15.33 & \second{0.5551} & \third{0.4244} \\
    PCRP-video~\citep{careagaRelighting}
    & \third{20.78} & 0.5852 & \third{0.3099} & 0.9838 & 0.9507 & \second{17.37} & \third{0.5812} & \second{0.2839} & \second{18.57} & \third{0.5214} & \second{0.2257} \\
    PBR~\citep{jakob2022mitsuba3}
    & 19.59 & \third{0.7111} & 0.3106 & 0.9755 & 0.9223 & 13.59 & 0.5130 & 0.3806 & \third{17.75} & 0.5082 & 0.4768 \\
\textbf{\OurMethod}
    & \best{24.16} & \best{0.7366} & \best{0.2096} & \best{0.9892} & \best{0.9598} & \best{20.27} & \best{0.8205} & \best{0.2114} & \best{19.55} & \best{0.6695} & \best{0.2030} \\
        \bottomrule
    \end{tabular}%
    }
     \vspace{-7pt}
     \caption{\textbf{Synthetic and Real-World Relighting Comparison.} \OurMethod accurately relights synthetic and real-world scenes (MIT, DL3DV) with better temporal consistency than the baselines.}
      \label{tab:main_relighting}
       \vspace{-10pt}
\end{table*}
\squeeze{-5mm}

\begin{figure*}[t]
    \centering
    \includegraphics[width=\textwidth,clip=true,trim=0mm 78mm 1mm 0mm]{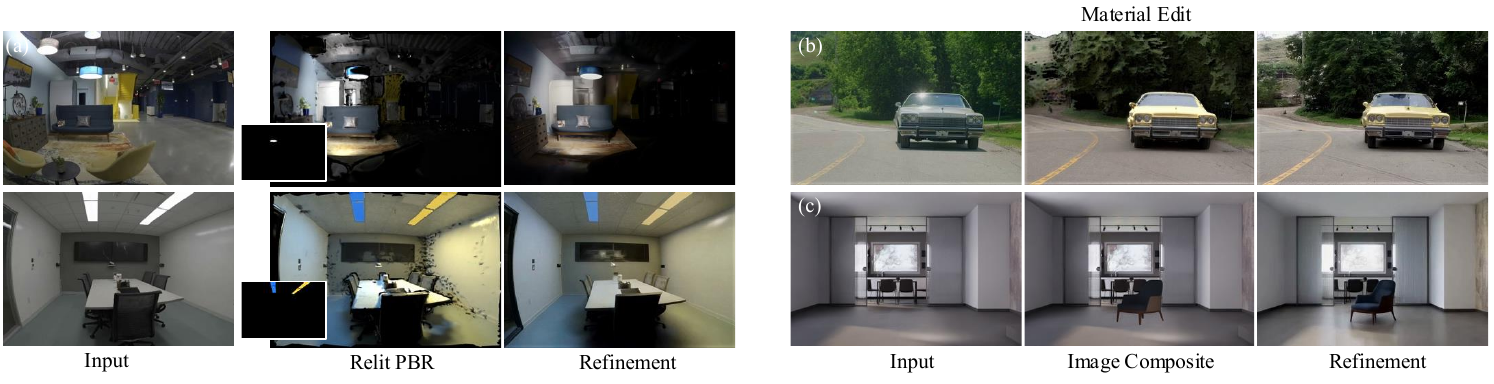}
     \vspace{-18pt}
    \caption{\textbf{Scene-Editing Applications.} \OurMethod's underlying explicit controllability allows for downstream scene editing applications, such as modifying lighting, materials, and inserting objects.
    }
        \vspace{-5pt}
    \label{fig:applications}
\end{figure*}

\begin{table*}[t]
    \centering
    \small
    \setlength{\tabcolsep}{2.7pt}

    \resizebox{\textwidth}{!}{%
    \begin{tabular}{@{}p{2.8cm}cccccccc@{}}
        \toprule
        & \multicolumn{5}{c}{Synthetic} & \multicolumn{3}{c}{Real-World \citep{murmann19, ling2024dl3dv}}
        \\
        \cmidrule(lr){2-6} \cmidrule(lr){7-9}
        Method & PSNR $\uparrow$ & SSIM $\uparrow$ & LPIPS $\downarrow$ & T-CLIP $\uparrow$ & Warp-SSIM $\uparrow$ & PSNR $\uparrow$ & SSIM $\uparrow$ & LPIPS $\downarrow$ \\
        \midrule

       Ours (No Real)
& \second{23.98} & \second{0.7324} & \best{0.2094} & \second{0.9878} & \best{0.9601} & \third{16.48} & \third{0.5515} & \third{0.3368} \\

Ours (No Synthetic)
& \third{15.86} & \third{0.5378} & \third{0.4041} & \third{0.9862} & \third{0.9488} & \second{16.76} & \second{0.5613} & \second{0.3048} \\

Ours (Full)
& \best{24.16} & \best{0.7366} & \second{0.2096} & \best{0.9892} & \second{0.9598} & \best{19.91} & \best{0.7450} & \best{0.2072} \\
        
        \bottomrule
    \end{tabular}%
     \vspace{-18pt}
    }
     \caption{\textbf{Effects of Ablating Synthetic or Real-World Data from Training.} Real-world metrics average MIT and DL3DV; an ablation without MIT data is in Appendix~\ref{supp:additional}.}
     \label{tab:ablation_relighting}
         \vspace{0pt}
\end{table*}

\begin{figure*}[t!]
    \centering
   \includegraphics[width=0.9\textwidth,clip=true,trim=0 71mm 31mm 0mm]{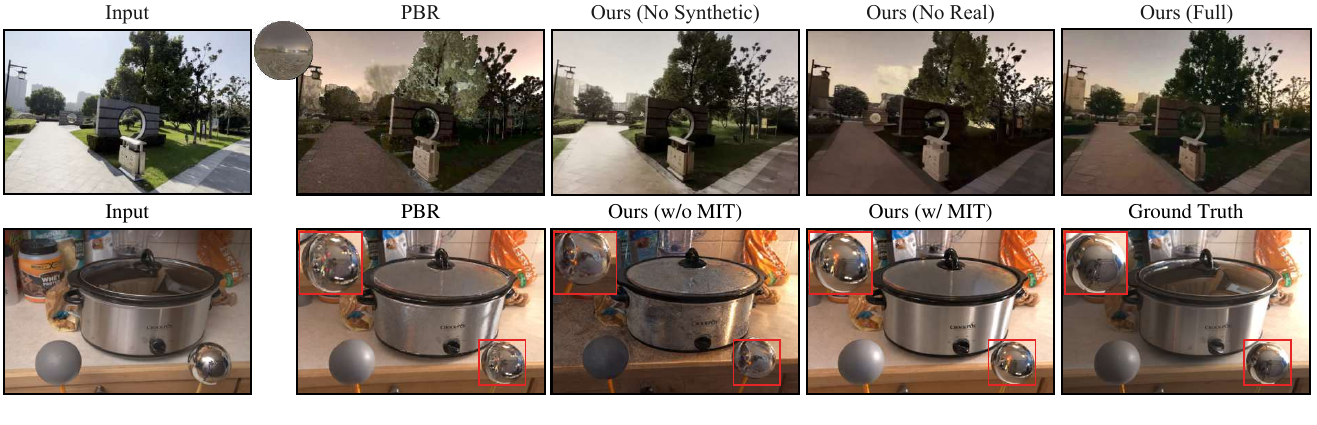}
    \vspace{-15pt}
    \caption{\textbf{Training Data Ablation Studies.} \emph{Top:} removing real-world data from training leads to unrealistic structures and hallucinations in the relighting, while removing synthetic data leads to poor relighting accuracy. Training with both data types leads to accurate and faithful relighting. \revision{\emph{Bottom:} a held-out MIT scene. Without MIT pairs the refiner flattens the brushed-steel pot and chrome ball into matte surfaces; training with MIT preserves metallicness and glare (insets: chrome ball).}
    }
        \vspace{15pt}

    \label{fig:ablations}
\end{figure*}
\subsection{Long-Form Consistent Relighting}
We evaluate \OurMethod's relighting consistency on long video sequences against prior methods on synthetic data. In Fig.~\ref{fig:prior_limitations}, We show qualitative results on 200-frame synthetic sequences with complex camera motion and dynamic content, comparing against end-to-end and two-stage baselines. \OurMethod maintains stable shadows, highlights, and global appearance throughout long sequences where baselines exhibit visible degradation. End-to-end methods suffer from chunked processing and temporal drift due to fixed context lengths and ungrounded illumination conditioning, while two-stage methods accumulate geometry and material estimation errors that produce flickering and localized relighting failures. Our formulation sidesteps these failures because the PBR rendering encodes target illumination, applying the same lighting condition uniformly across all frames regardless of sequence length, while the diffusion model refines the video and preserves this consistent structure.
 \vspace{-1pt}
\paragraph{User study.}
Reconstruction metrics do not capture everything a viewer notices, so we also conduct a user study (protocol in Appendix~\ref{supp:userstudy}). Each of 16 participants viewed 11 comparison sets, each pairing a source video and its target illumination with five relit results (ours and the four baselines) in randomized order, and picked the single best result under four criteria: temporal consistency, relighting accuracy, overall visual quality, and scene-content preservation. \OurMethod{} was the most preferred method on every criterion, with the widest margin on relighting accuracy (Table~\ref{tab:user_study}).


 \squeeze{-2mm}
\subsection{Editing Applications}
\label{sec:scene_editing}

The generative PBR rendering refinement formulation naturally supports downstream editing tasks beyond direct relighting. Fig.~\ref{fig:applications} demonstrates three applications. For light insertion, we manually place a virtual light source with specified intensity and direction, and forward render. For material editing, we modify the estimated albedo or material buffers before re-rendering. For object insertion, we composite a rendered object into the PBR video and let the refiner harmonize the result. 
In all cases, edits are applied with the forward renderer and refined by the same finetuned model, requiring no additional training.

\subsection{Ablation Studies}

The quantitative impact of ablating different training data sources is presented in Table~\ref{tab:ablation_relighting}, with qualitative results shown in Fig.~\ref{fig:ablations}. Training on synthetic data only \revision{preserves synthetic relighting accuracy but degrades real-world performance, as} the refiner fails to correct complex structures such as foliage and thin geometry, and can hallucinate artifacts. Training only on real-world data produces photorealistic refinement but poor relighting control. Training on both data types yields accurate lighting control and photorealistic refinement across diverse structures. The artifact-matched PBR renders are critical; by exposing the model to realistic rendering artifacts during training, it learns to correct in-the-wild artifacts at inference.

\vspace{-15pt}
\subsection{Limitations}
\label{sec:limitations}

Our inverse rendering remains noisy on specular, transparent, or metallic surfaces and on scenes with weak parallax, motion blur, or thin structures; such errors propagate to the PBR proxy and can exceed what the refiner corrects. \revision{We discuss these limitations, promising remedies, and failure cases in Appendix~\ref{supp:failures}.}
\vspace{-10pt}

\section{Conclusion}
We presented \OurMethod, a diffusion relighting framework that reformulates video relighting as PBR refinement. Our key insight is that a forward-rendered PBR proxy already captures most structured lighting changes and provides temporal consistency by construction, leaving the diffusion model to focus on artifact correction rather than global relighting. This division of tasks yields higher fidelity lighting control, stable long-form relighting, and natural extension to scene editing. We further presented a novel data curation pipeline for artifact-matched supervision for both synthetic and real-world scenes to bridge the gap between training and generalization at inference. Experimental results show that \OurMethod achieves state-of-the-art relighting performance on complex synthetic and real-world scenes while enabling consistent relighting over arbitrarily long videos.

{\revstart
\subsubsection*{Ethics statement}
\OurMethod{} enables realistic video relighting and scene editing, with potential applications in content creation and visual effects. These capabilities could also be misused to create deceptive footage by altering the apparent time of day, illumination, or environmental conditions of a recorded event. We encourage clearly identifying relit outputs as edited content and pairing their distribution with provenance and content-authentication mechanisms, such as watermarking. Edited videos should not be presented as unaltered recordings or evidence of real-world events.

Our user study collected anonymous preference judgments from volunteer participants without collecting personal data. The study assessed perceived visual quality, lighting, and temporal consistency; these judgments do not establish the factual authenticity or physical accuracy of the generated outputs.

Our experiments use publicly available resources, including Objaverse, Poly Haven, DL3DV-10K, and MIT Multi-Illumination, under their respective licenses. In-the-wild examples additionally include stock footage and AI-generated videos. These generated examples are used to evaluate relighting behavior and are not presented as recordings of real events. Any redistribution of source assets or derived data should respect the applicable licenses, terms of use, and privacy constraints.

\subsubsection*{Reproducibility statement}
Sec.~\ref{sec:method_start} and Appendix~\ref{supp:artifact_matched_data} describe the full data construction, including scene, clip, and split counts (Table~\ref{tab:data_split}); Appendix~\ref{supp:refiner} gives the refiner, conditioning noise, inference settings, and the complete long-video fusion procedure; Appendix~\ref{supp:runtime} reports per-component runtime and memory; and Appendix~\ref{supp:baselines} details the baseline settings, including the PCRP-Video implementation. We will release our dataset, benchmark, metrics, and code.
\par}

\bibliography{main}

\begin{thebibliography}{40}
\providecommand{\natexlab}[1]{#1}
\providecommand{\url}[1]{\texttt{#1}}
\expandafter\ifx\csname urlstyle\endcsname\relax
  \providecommand{\doi}[1]{doi: #1}\else
  \providecommand{\doi}{doi: \begingroup \urlstyle{rm}\Url}\fi

\bibitem[Bar-Tal et~al.(2023)Bar-Tal, Yariv, Lipman, and
  Dekel]{bar2023multidiffusion}
Omer Bar-Tal, Lior Yariv, Yaron Lipman, and Tali Dekel.
\newblock {MultiDiffusion}: Fusing diffusion paths for controlled image
  generation.
\newblock In \emph{ICML}, 2023.

\bibitem[Bi et~al.(2024)Bi, Zeng, Zeng, Pei, Feng, Zhou, and Wu]{bi2024gs3}
Zoubin Bi, Yixin Zeng, Chong Zeng, Fan Pei, Xiang Feng, Kun Zhou, and Hongzhi
  Wu.
\newblock Gs3: Efficient relighting with triple gaussian splatting.
\newblock In \emph{SIGGRAPH Asia}, 2024.

\bibitem[Blattmann et~al.(2023)Blattmann, Dockhorn, Kulal, Mendelevitch,
  Kilian, Lorenz, Levi, English, Voleti, Letts, Jampani, and
  Rombach]{blattmann2023stablevideodiffusionscaling}
Andreas Blattmann, Tim Dockhorn, Sumith Kulal, Daniel Mendelevitch, Maciej
  Kilian, Dominik Lorenz, Yam Levi, Zion English, Vikram Voleti, Adam Letts,
  Varun Jampani, and Robin Rombach.
\newblock Stable video diffusion: Scaling latent video diffusion models to
  large datasets.
\newblock \emph{arXiv preprint arXiv:2311.15127}, 2023.

\bibitem[Burley(2012)]{burley2012physically}
Brent Burley.
\newblock Physically-based shading at disney.
\newblock In \emph{SIGGRAPH Courses}, 2012.

\bibitem[Careaga \& Aksoy(2025)Careaga and Aksoy]{careagaRelighting}
Chris Careaga and Ya\u{g}{\i}z Aksoy.
\newblock Physically controllable relighting of photographs.
\newblock In \emph{SIGGRAPH}, 2025.

\bibitem[Chen et~al.(2025)Chen, Lin, and Zhang]{chen2025gigs}
Hongze Chen, Zehong Lin, and Jun Zhang.
\newblock Gi-gs: Global illumination decomposition on gaussian splatting for
  inverse rendering.
\newblock In \emph{ICLR}, 2025.

\bibitem[Cook \& Torrance(1982)Cook and Torrance]{cook1982reflectance}
Robert~L. Cook and Kenneth~E. Torrance.
\newblock A reflectance model for computer graphics.
\newblock \emph{ACM TOG}, 1982.

\bibitem[Deitke et~al.(2022)Deitke, Schwenk, Salvador, Weihs, Michel,
  VanderBilt, Schmidt, Ehsani, Kembhavi, and Farhadi]{objaverse}
Matt Deitke, Dustin Schwenk, Jordi Salvador, Luca Weihs, Oscar Michel, Eli
  VanderBilt, Ludwig Schmidt, Kiana Ehsani, Aniruddha Kembhavi, and Ali
  Farhadi.
\newblock Objaverse: A universe of annotated 3d objects.
\newblock \emph{arXiv preprint arXiv:2212.08051}, 2022.

\bibitem[Demes(2024)]{cctexturekey}
Lennart Demes.
\newblock {ambientCG} (formerly {CC0 Textures}): Free public domain textures,
  {HDRIs} and models, 2024.
\newblock https://ambientcg.com/.

\bibitem[Fischer et~al.(2025)Fischer, Bul{\`o}, Yang, Keetha, Porzi, M\"uller,
  Schwarz, Luiten, Pollefeys, and Kontschieder]{fischer2025flowr}
Tobias Fischer, Samuel~Rota Bul{\`o}, Yung-Hsu Yang, Nikhil Keetha, Lorenzo
  Porzi, Norman M\"uller, Katja Schwarz, Jonathon Luiten, Marc Pollefeys, and
  Peter Kontschieder.
\newblock {FlowR}: Flowing from sparse to dense 3d reconstructions.
\newblock In \emph{ICCV}, 2025.

\bibitem[He et~al.(2025)He, Liang, Munkberg, Hasselgren, Vijaykumar, Keller,
  Fidler, Gilitschenski, Gojcic, and Wang]{he2025unirelight}
Kai He, Ruofan Liang, Jacob Munkberg, Jon Hasselgren, Nandita Vijaykumar,
  Alexander Keller, Sanja Fidler, Igor Gilitschenski, Zan Gojcic, and Zian
  Wang.
\newblock Unirelight: Learning joint decomposition and synthesis for video
  relighting.
\newblock In \emph{NeurIPS}, 2025.

\bibitem[Jakob et~al.(2022)Jakob, Speierer, Roussel, Nimier-David, Vicini,
  Zeltner, Nicolet, Crespo, Leroy, and Zhang]{jakob2022mitsuba3}
Wenzel Jakob, S\'ebastien Speierer, Nicolas Roussel, Merlin Nimier-David, Delio
  Vicini, Tizian Zeltner, Baptiste Nicolet, Miguel Crespo, Vincent Leroy, and
  Ziyi Zhang.
\newblock Mitsuba 3 renderer, 2022.
\newblock https://mitsuba-renderer.org.

\bibitem[Jin et~al.(2023)Jin, Liu, Xu, Zhang, Han, Bi, Zhou, Xu, and
  Su]{jin2023tensoir}
Haian Jin, Isabella Liu, Peijia Xu, Xiaoshuai Zhang, Songfang Han, Sai Bi,
  Xiaowei Zhou, Zexiang Xu, and Hao Su.
\newblock Tensoir: Tensorial inverse rendering.
\newblock In \emph{CVPR}, 2023.

\bibitem[Karis(2013)]{karis2013real}
Brian Karis.
\newblock Real shading in unreal engine 4.
\newblock In \emph{SIGGRAPH Courses}, 2013.

\bibitem[Li et~al.(2024)Li, Tucker, Cole, Wang, Jin, Ye, Kanazawa, Holynski,
  and Snavely]{li2024megasamaccuratefastrobust}
Zhengqi Li, Richard Tucker, Forrester Cole, Qianqian Wang, Linyi Jin, Vickie
  Ye, Angjoo Kanazawa, Aleksander Holynski, and Noah Snavely.
\newblock Megasam: Accurate, fast, and robust structure and motion from casual
  dynamic videos.
\newblock \emph{arXiv preprint arXiv:2412.04463}, 2024.

\bibitem[Liang et~al.(2025)Liang, Gojcic, Ling, Munkberg, Hasselgren, Lin, Gao,
  Keller, Vijaykumar, Fidler, and Wang]{liang2025diffusionrenderer}
Ruofan Liang, Zan Gojcic, Huan Ling, Jacob Munkberg, Jon Hasselgren, Chih-Hao
  Lin, Jun Gao, Alexander Keller, Nandita Vijaykumar, Sanja Fidler, and Zian
  Wang.
\newblock Diffusion renderer: Neural inverse and forward rendering with video
  diffusion models.
\newblock In \emph{CVPR}, 2025.

\bibitem[Liang et~al.(2026)Liang, M{\"u}ller, Weber, Zauss, Vijaykumar,
  Kontschieder, and Richardt]{liang2026luxremix}
Ruofan Liang, Norman M{\"u}ller, Ethan Weber, Duncan Zauss, Nandita Vijaykumar,
  Peter Kontschieder, and Christian Richardt.
\newblock {LuxRemix}: Lighting decomposition and remixing for indoor scenes.
\newblock \emph{arXiv preprint arXiv:2601.15283}, 2026.

\bibitem[Ling et~al.(2024)Ling, Sheng, Tu, Zhao, Xin, Wan, Yu, Guo, Yu, Lu,
  et~al.]{ling2024dl3dv}
Lu~Ling, Yichen Sheng, Zhi Tu, Wentian Zhao, Cheng Xin, Kun Wan, Lantao Yu,
  Qianyu Guo, Zixun Yu, Yawen Lu, et~al.
\newblock Dl3dv-10k: A large-scale scene dataset for deep learning-based 3d
  vision.
\newblock In \emph{CVPR}, 2024.

\bibitem[Litman et~al.(2025)Litman, De~la Torre, and
  Tulsiani]{litman2025lightswitch}
Yehonathan Litman, Fernando De~la Torre, and Shubham Tulsiani.
\newblock Lightswitch: Multi-view relighting with material-guided diffusion.
\newblock In \emph{ICCV}, 2025.

\bibitem[Liu et~al.(2025)Liu, Yuan, Dong, Xing, Wang, Zhao, Chen, and
  Wang]{UniLumos}
Pengwei Liu, Hangjie Yuan, Bo~Dong, Jiazheng Xing, Jinwang Wang, Rui Zhao,
  Weihua Chen, and Fan Wang.
\newblock Unilumos: Fast and unified image and video relighting with
  physics-plausible feedback.
\newblock In \emph{NeurIPS}, 2025.

\bibitem[Liu et~al.(2026)Liu, Chen, Huang, Xu, Zhang, Ye, Li, Cao, Li, Zhao,
  and Liu]{liu2025lightx}
Tianqi Liu, Zhaoxi Chen, Zihao Huang, Shaocong Xu, Saining Zhang, Chongjie Ye,
  Bohan Li, Zhiguo Cao, Wei Li, Hao Zhao, and Ziwei Liu.
\newblock Light-x: Generative 4d video rendering with camera and illumination
  control.
\newblock In \emph{ICLR}, 2026.

\bibitem[Loshchilov \& Hutter(2017)Loshchilov and
  Hutter]{loshchilov2017decoupled}
Ilya Loshchilov and Frank Hutter.
\newblock Decoupled weight decay regularization.
\newblock \emph{arXiv preprint arXiv:1711.05101}, 2017.

\bibitem[Murmann et~al.(2019)Murmann, Gharbi, Aittala, and Durand]{murmann19}
Lukas Murmann, Michael Gharbi, Miika Aittala, and Fredo Durand.
\newblock A dataset of multi-illumination images in the wild.
\newblock In \emph{ICCV}, 2019.

\bibitem[Phongthawee et~al.(2024)Phongthawee, Chinchuthakun, Sinsunthithet,
  Jampani, Raj, Khungurn, and Suwajanakorn]{phongthawee2024diffusionlight}
Pakkapon Phongthawee, Worameth Chinchuthakun, Nontaphat Sinsunthithet, Varun
  Jampani, Amit Raj, Pramook Khungurn, and Supasorn Suwajanakorn.
\newblock Diffusionlight: Light probes for free by painting a chrome ball.
\newblock In \emph{CVPR}, 2024.

\bibitem[Walter et~al.(2007)Walter, Marschner, Li, and
  Torrance]{walter2007microfacet}
Bruce Walter, Stephen~R. Marschner, Hongsong Li, and Kenneth~E. Torrance.
\newblock Microfacet models for refraction through rough surfaces.
\newblock In \emph{EGSR}, 2007.

\bibitem[Wang et~al.(2004)Wang, Bovik, Sheikh, and Simoncelli]{Wang2004SSIM}
Zhou Wang, Alan~C Bovik, Hamid~R Sheikh, and Eero~P Simoncelli.
\newblock Image quality assessment: From error visibility to structural
  similarity.
\newblock \emph{IEEE TIP}, 2004.

\bibitem[Wu et~al.(2025)Wu, Zhang, Turki, Ren, Gao, Shou, Fidler, Gojcic, and
  Ling]{wu2025difix3dplus}
Jay~Zhangjie Wu, Yuxuan Zhang, Haithem Turki, Xuanchi Ren, Jun Gao, Mike~Zheng
  Shou, Sanja Fidler, Zan Gojcic, and Huan Ling.
\newblock Difix3d+: Improving 3d reconstructions with single-step diffusion
  models.
\newblock In \emph{CVPR}, 2025.

\bibitem[Xing et~al.(2026{\natexlab{a}})Xing, Du, Yuan, Liu, Xu, Ci, Niu, Chen,
  Wang, and Liu]{LumosX}
Jiazheng Xing, Fei Du, Hangjie Yuan, Pengwei Liu, Hongbin Xu, Hai Ci, Ruigang
  Niu, Weihua Chen, Fan Wang, and Yong Liu.
\newblock Lumosx: Relate any identities with their attributes for personalized
  video generation.
\newblock In \emph{ICLR}, 2026{\natexlab{a}}.

\bibitem[Xing et~al.(2026{\natexlab{b}})Xing, Henzler, Hur, Li, Barron,
  Srinivasan, and Verbin]{xing2026gr3en}
Xiaoyan Xing, Philipp Henzler, Junhwa Hur, Runze Li, Jonathan~T. Barron,
  Pratul~P. Srinivasan, and Dor Verbin.
\newblock Gr3en: Generative relighting for 3d environments.
\newblock In \emph{SIGGRAPH}, 2026{\natexlab{b}}.

\bibitem[Yang et~al.(2025)Yang, Teng, Zheng, Ding, Huang, Xu, Yang, Hong,
  Zhang, Feng, Yin, Zhang, Wang, Cheng, Xu, Gu, Dong, and
  Tang]{yang2025cogvideoxtexttovideodiffusionmodels}
Zhuoyi Yang, Jiayan Teng, Wendi Zheng, Ming Ding, Shiyu Huang, Jiazheng Xu,
  Yuanming Yang, Wenyi Hong, Xiaohan Zhang, Guanyu Feng, Da~Yin, Yuxuan Zhang,
  Weihan Wang, Yean Cheng, Bin Xu, Xiaotao Gu, Yuxiao Dong, and Jie Tang.
\newblock {CogVideoX}: Text-to-video diffusion models with an expert
  transformer.
\newblock In \emph{ICLR}, 2025.

\bibitem[Yao et~al.(2022)Yao, Zhang, Liu, Qu, Fang, McKinnon, Tsin, and
  Quan]{yao2022neilf}
Yao Yao, Jingyang Zhang, Jingbo Liu, Yihang Qu, Tian Fang, David McKinnon,
  Yanghai Tsin, and Long Quan.
\newblock Neilf: Neural incident light field for physically-based material
  estimation.
\newblock In \emph{ECCV}, 2022.

\bibitem[Zaal et~al.(2024)]{polyhaven}
Greg Zaal et~al.
\newblock Poly haven: The public 3d asset library, 2024.
\newblock https://polyhaven.com/.

\bibitem[Zeng et~al.(2024)Zeng, Dong, Peers, Kong, Wu, and
  Tong]{zeng2024dilightnet}
Chong Zeng, Yue Dong, Pieter Peers, Youkang Kong, Hongzhi Wu, and Xin Tong.
\newblock Dilightnet: Fine-grained lighting control for diffusion-based image
  generation.
\newblock In \emph{SIGGRAPH}, 2024.

\bibitem[Zeng et~al.(2025)Zeng, Liu, Feng, Miao, Gao, Qu, Zhang, Wang, and
  Yuan]{zeng2025lumen}
Jianshu Zeng, Yuxuan Liu, Yutong Feng, Chenxuan Miao, Zixiang Gao, Jiwang Qu,
  Jianzhang Zhang, Bin Wang, and Kun Yuan.
\newblock Lumen: Consistent video relighting and harmonious background
  replacement with video generative models.
\newblock \emph{arXiv preprint arXiv:2508.12945}, 2025.

\bibitem[Zhang et~al.(2023)Zhang, Yao, Li, Liu, Fang, McKinnon, Tsin, and
  Quan]{zhang2023neilfpp}
Jingyang Zhang, Yao Yao, Shiwei Li, Jingbo Liu, Tian Fang, David McKinnon,
  Yanghai Tsin, and Long Quan.
\newblock Neilf++: Inter-reflectable light fields for geometry and material
  estimation.
\newblock In \emph{ICCV}, 2023.

\bibitem[Zhang et~al.(2018)Zhang, Isola, Efros, Shechtman, and
  Wang]{Zhang2018LPIPS}
Richard Zhang, Phillip Isola, Alexei~A Efros, Eli Shechtman, and Oliver Wang.
\newblock The unreasonable effectiveness of deep features as a perceptual
  metric.
\newblock In \emph{CVPR}, 2018.

\bibitem[Zhang et~al.(2021)Zhang, Srinivasan, Deng, Debevec, Freeman, and
  Barron]{zhang2021nerfactor}
Xiuming Zhang, Pratul~P. Srinivasan, Boyang Deng, Paul Debevec, William~T.
  Freeman, and Jonathan~T. Barron.
\newblock Nerfactor: Neural factorization of shape and reflectance under an
  unknown illumination.
\newblock \emph{ACM TOG}, 2021.

\bibitem[Zhang et~al.(2026)Zhang, T{\'o}thov{\'a}, Wang, Yin, Turki, de~Lutio,
  Chang, Litany, Fidler, and Gojcic]{zhang2026diffusionharmonizer}
Yuxuan Zhang, Katar{\'i}na T{\'o}thov{\'a}, Zian Wang, Kangxue Yin, Haithem
  Turki, Riccardo de~Lutio, Yen-Yu Chang, Or~Litany, Sanja Fidler, and Zan
  Gojcic.
\newblock Diffusionharmonizer: Bridging neural reconstruction and
  photorealistic simulation with online diffusion enhancer.
\newblock \emph{arXiv preprint arXiv:2602.24096}, 2026.

\bibitem[Zhou et~al.(2025{\natexlab{a}})Zhou, Gao, Voleti, Vasishta, Yao, Boss,
  Torr, Rupprecht, and Jampani]{zhou2025stable}
Jensen Zhou, Hang Gao, Vikram Voleti, Aaryaman Vasishta, Chun-Han Yao, Mark
  Boss, Philip Torr, Christian Rupprecht, and Varun Jampani.
\newblock Stable virtual camera: Generative view synthesis with diffusion
  models.
\newblock In \emph{Proceedings of the IEEE/CVF International Conference on
  Computer Vision (ICCV)}, pp.\  12405--12414, 2025{\natexlab{a}}.

\bibitem[Zhou et~al.(2025{\natexlab{b}})Zhou, Bu, Ling, Zhang, Wu, Huang, Li,
  Dong, Zang, Cao, Rao, Wang, and Niu]{zhou2025lightavideo}
Yujie Zhou, Jiazi Bu, Pengyang Ling, Pan Zhang, Tong Wu, Qidong Huang, Jinsong
  Li, Xiaoyi Dong, Yuhang Zang, Yuhang Cao, Anyi Rao, Jiaqi Wang, and Li~Niu.
\newblock Light-a-video: Training-free video relighting via progressive light
  fusion.
\newblock In \emph{ICCV}, 2025{\natexlab{b}}.

\end{thebibliography}
\bibliographystyle{iclr2027_conference}

\newpage
\appendix

\section{Supplementary}

\label{sec:supp}

\begin{figure*}[t]
    \centering
    \includegraphics[width=\textwidth,clip=true,trim=0 0mm 0mm 0mm]{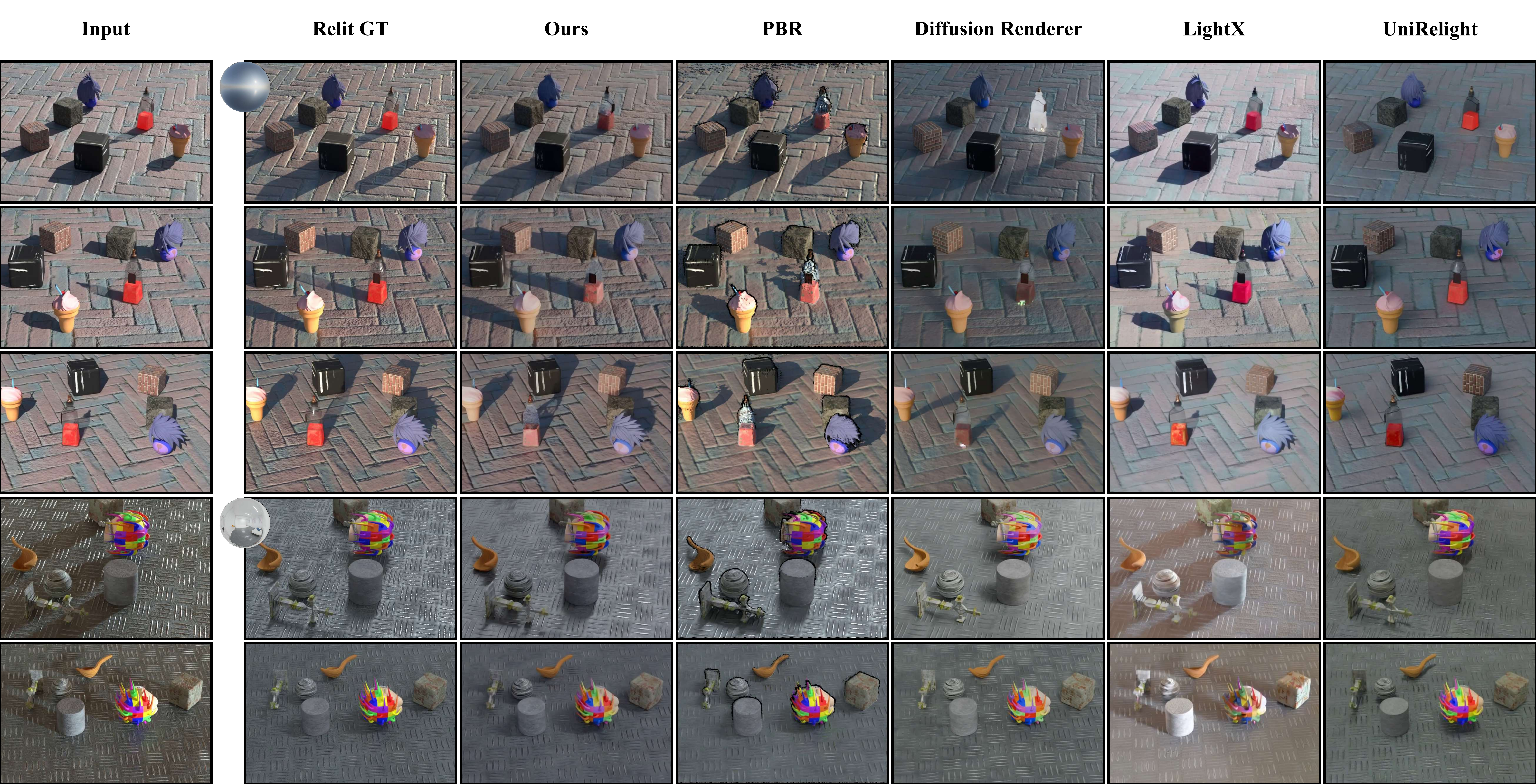}
    \includegraphics[width=\textwidth,clip=true,trim=0 0mm 0mm 0mm]{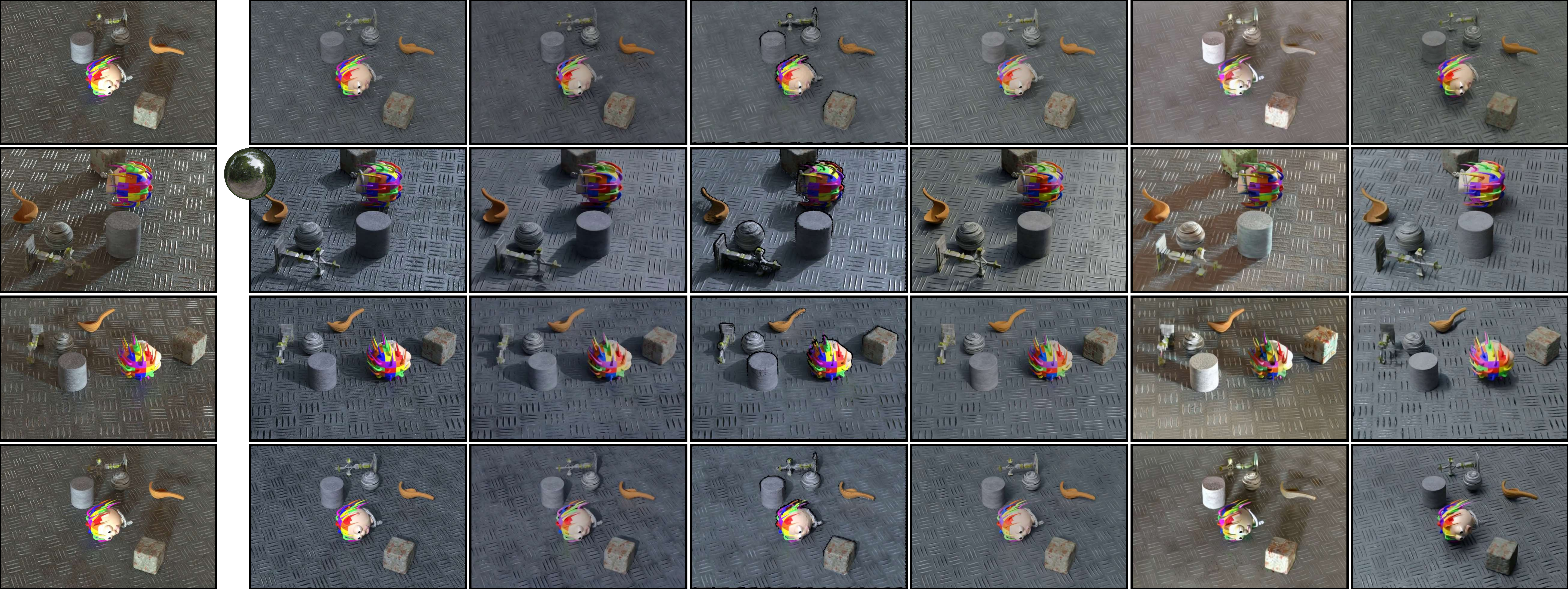}
    \squeeze{-5mm}
    \caption{Qualitative comparison on synthetic relighting videos with a camera orbiting a static scene. Each row shows a sampled frame under a target HDR environment map, and columns compare the input, relit ground truth, our result, PBR rendering, and prior image/video relighting baselines. The environment sphere visualizes the target illumination.
    }
    \label{fig:supp_setting1}
\end{figure*}

\begin{figure*}[t]
    \centering
    \includegraphics[width=\textwidth,clip=true,trim=0 0mm 0mm 0mm]{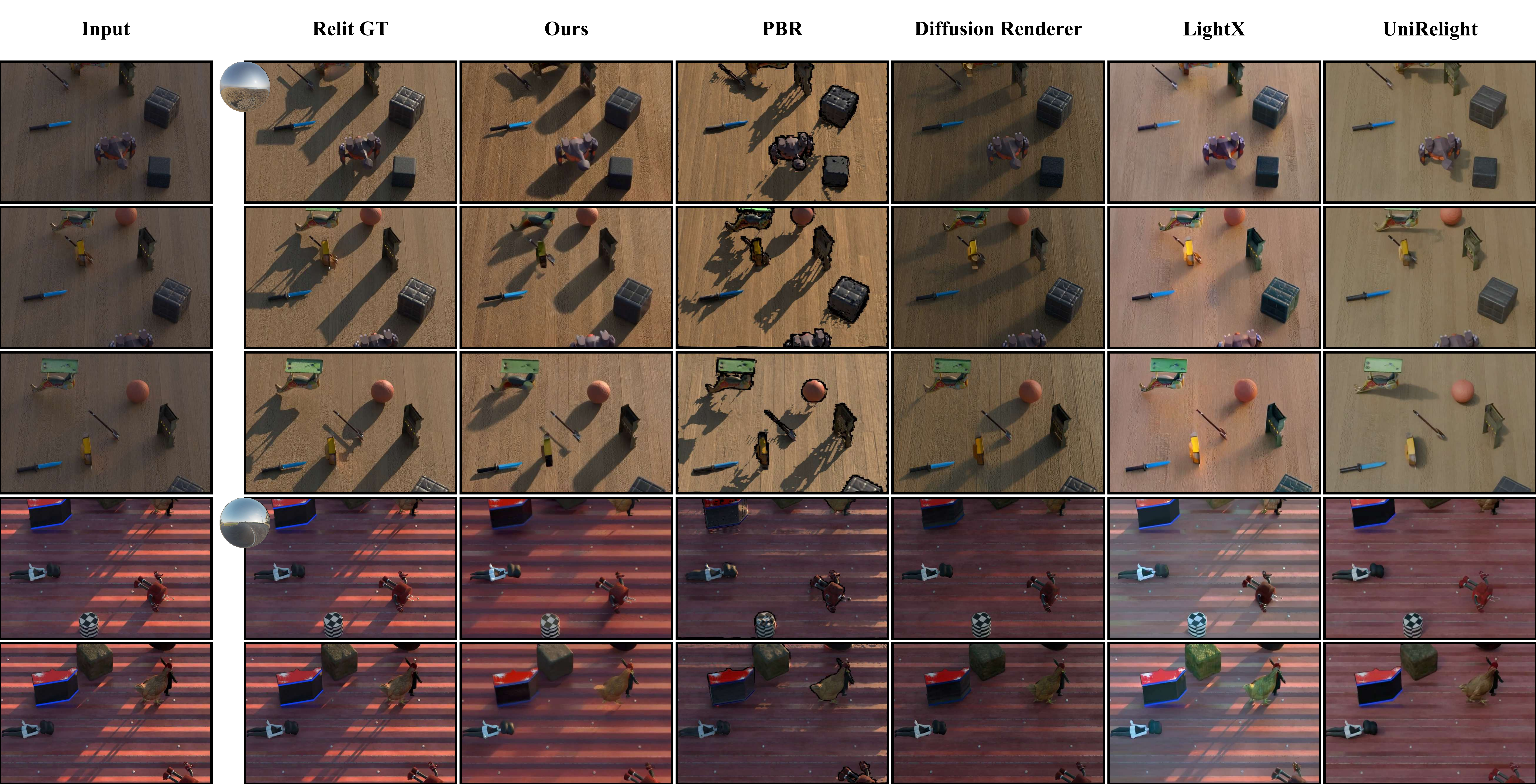}
    \includegraphics[width=\textwidth,clip=true,trim=0 0mm 0mm 0mm]{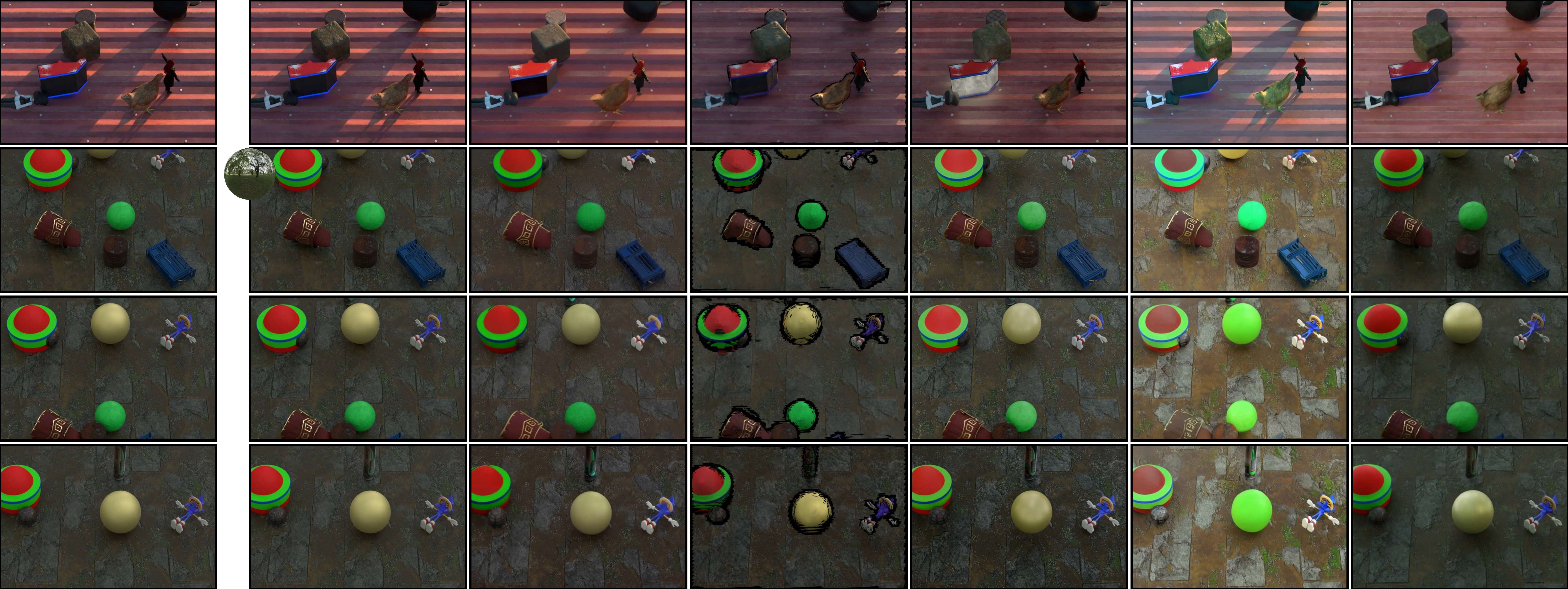}
    \squeeze{-5mm}
    \caption{Qualitative comparison on synthetic videos with moving cameras and dynamic scenes. This setting evaluates temporal relighting under simultaneous viewpoint and scene motion. Our method better preserves object geometry, shadow structure, and target-illumination consistency across frames compared with PBR rendering and existing relighting baselines.
    }
    \label{fig:supp_setting2}
\end{figure*}

\begin{figure*}[t]
    \centering
    \includegraphics[width=\textwidth,clip=true,trim=0 0mm 0mm 0mm]{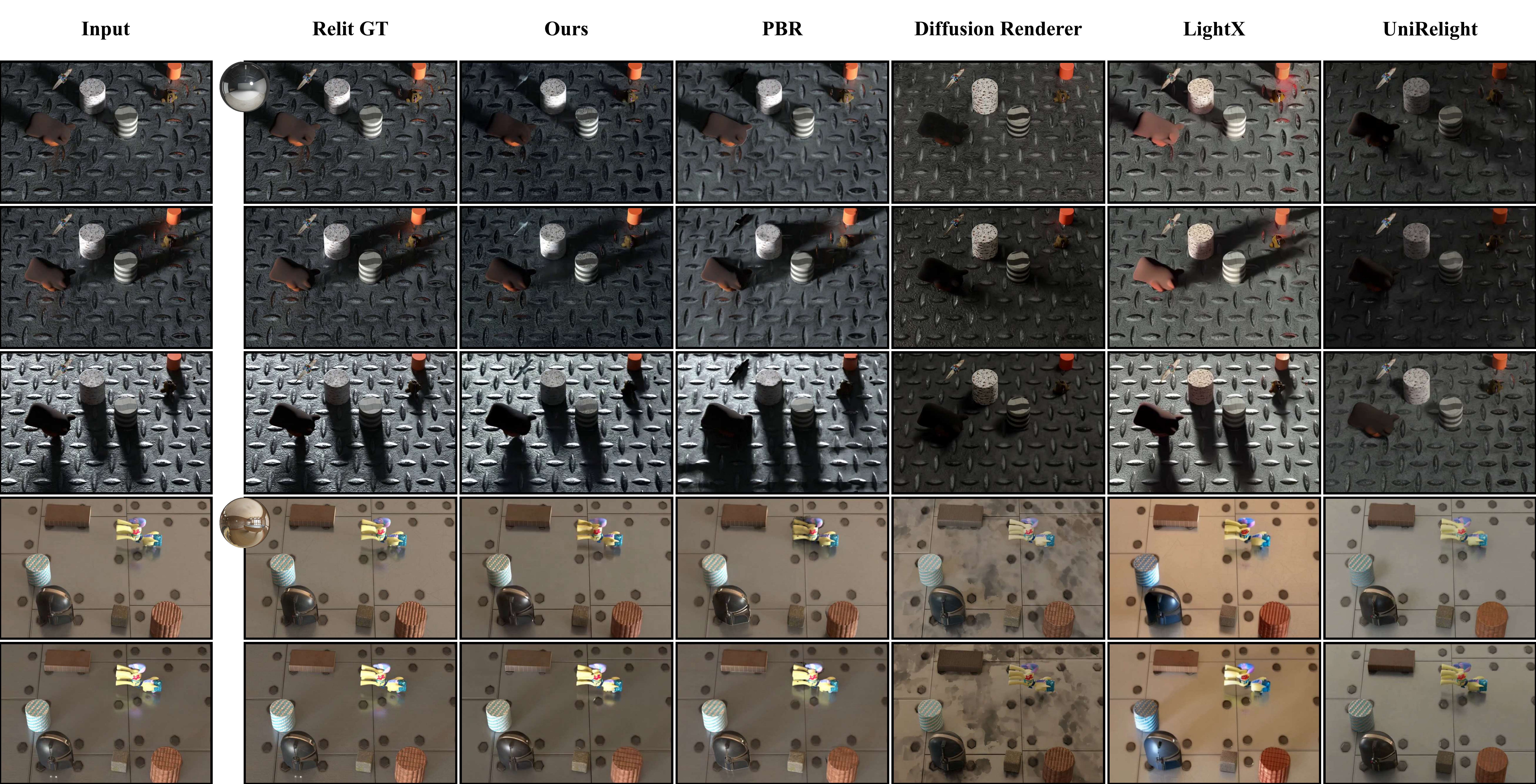}
    \includegraphics[width=\textwidth,clip=true,trim=0 0mm 0mm 0mm]{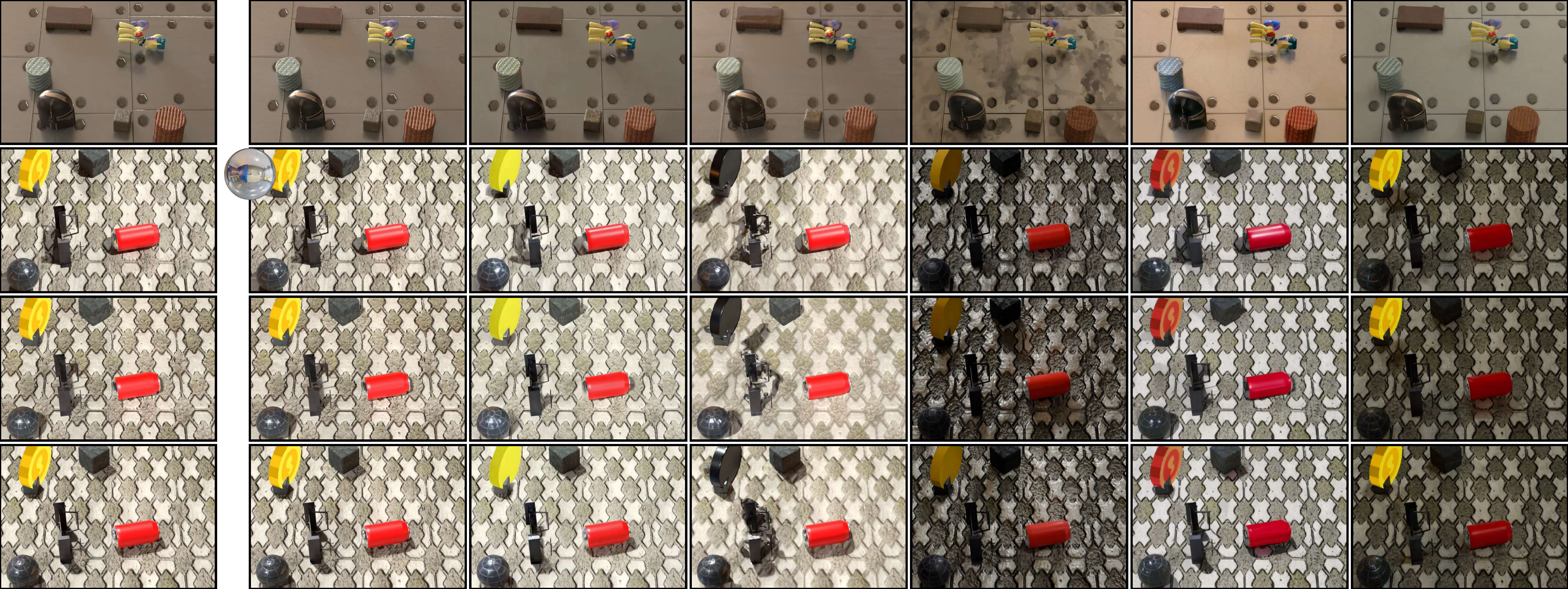}
    \squeeze{-5mm}
    \caption{Qualitative comparison on synthetic videos captured by a static camera under rotating target illumination. This setting isolates the effect of illumination change while keeping scene geometry fixed. Our method produces more coherent shading and highlights that follow the target environment map.
    }
    \label{fig:supp_setting3}
\end{figure*}

\begin{figure*}[t]
    \centering
    \includegraphics[width=\textwidth,clip=true,trim=0 0mm 0mm 0mm]{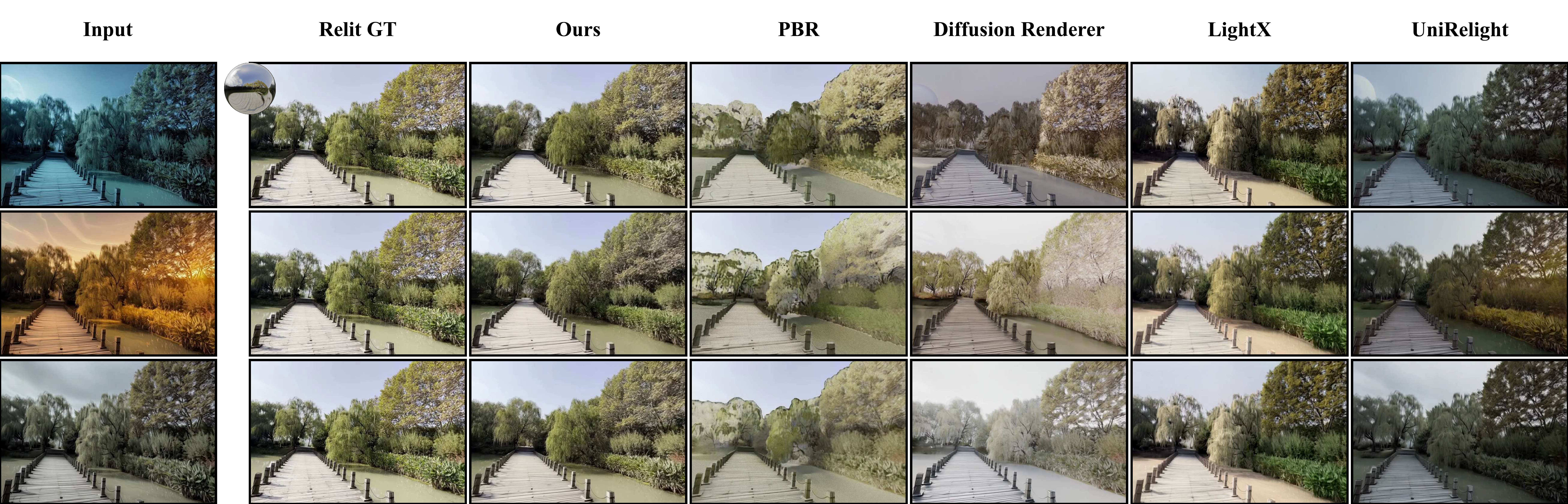}
    \includegraphics[width=\textwidth,clip=true,trim=0 0mm 0mm 0mm]{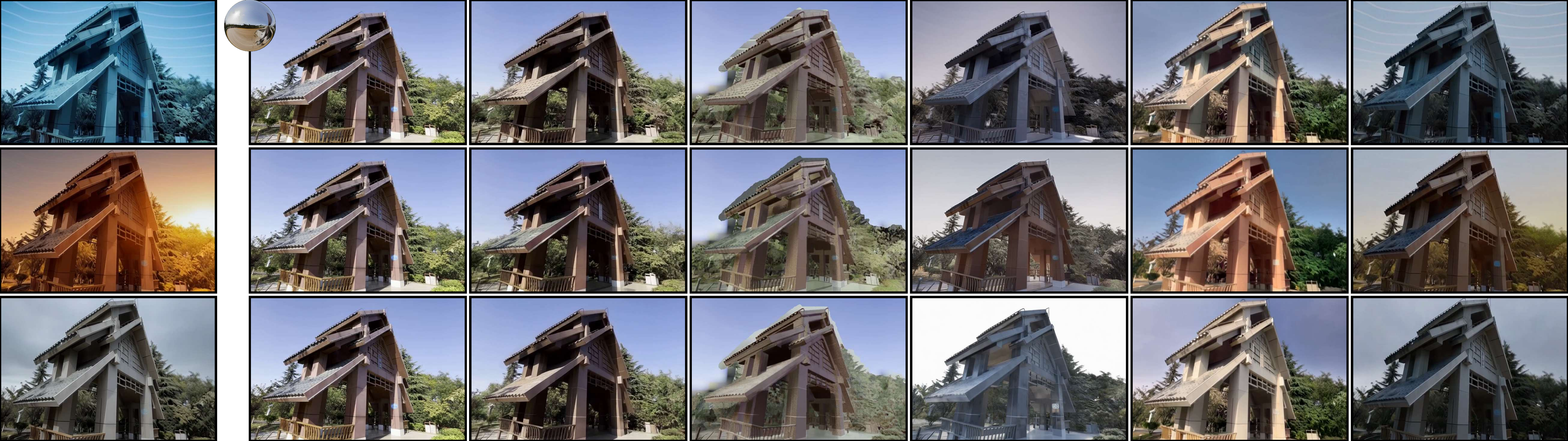}
    \includegraphics[width=\textwidth,clip=true,trim=0 0mm 0mm 0mm]{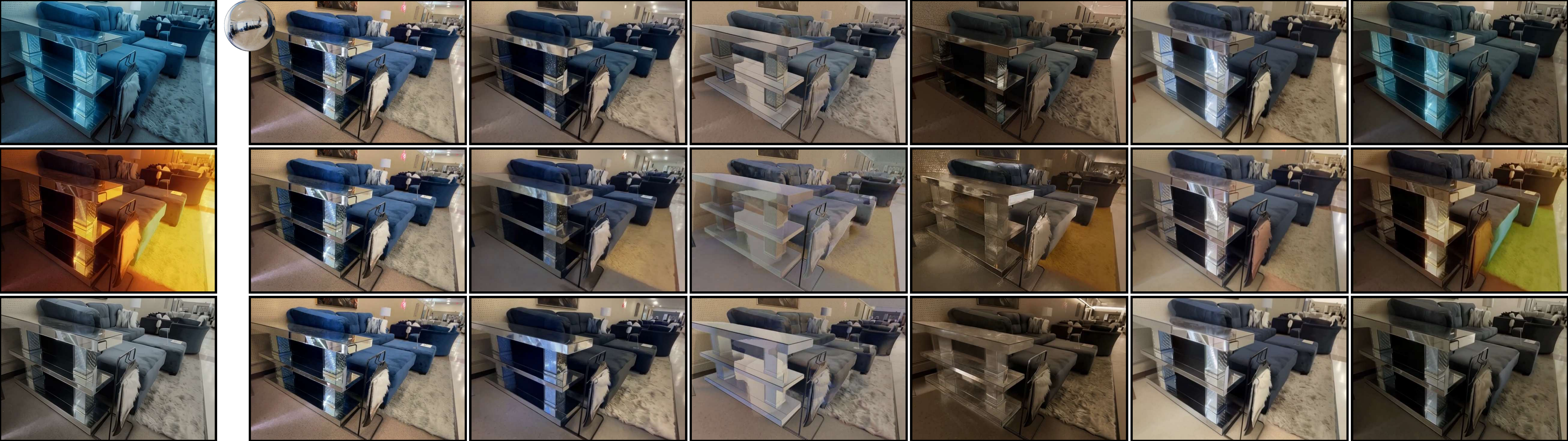}
    \squeeze{-5mm}
    \caption{Qualitative comparison on real-world video pseudo-pairs from DL3DV. We first relight a DL3DV video to a new environment using LaV, and then use this relit video as the input for all methods, with the task of relighting it back to the original video. This setting evaluates whether each method can recover the original illumination and appearance from a real video under altered lighting.
    }
    \label{fig:supp_setting4}
\end{figure*}

\begin{figure*}[t]
    \centering
    \includegraphics[width=\textwidth,clip=true,trim=0 0mm 0mm 0mm]{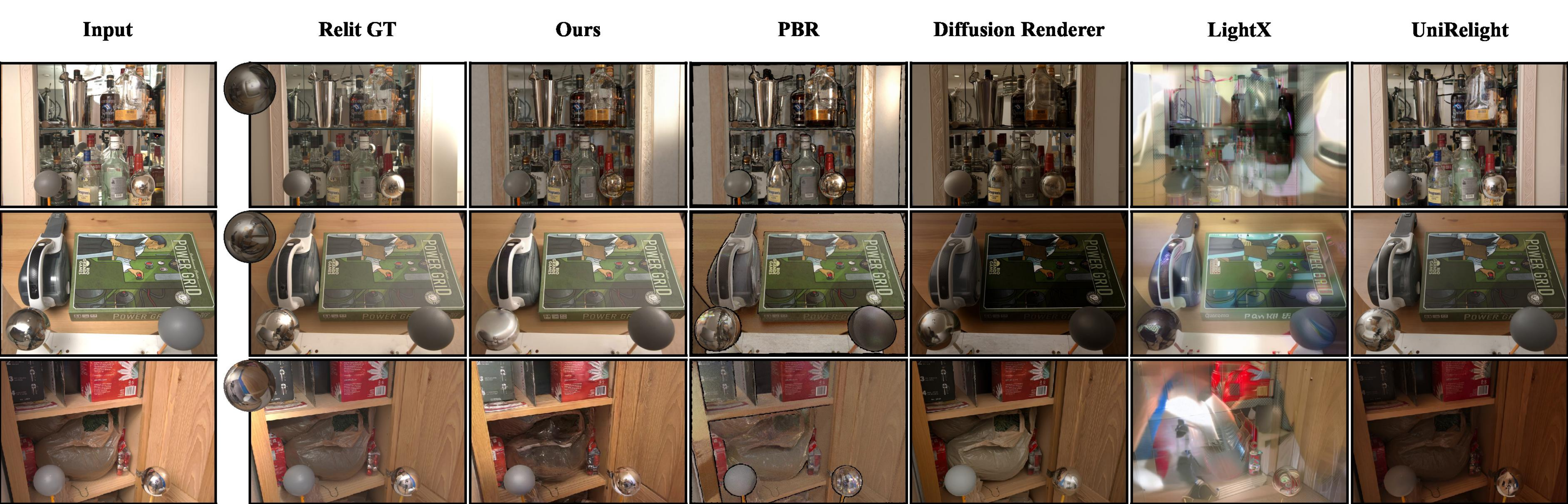}
    \squeeze{-5mm}
    \caption{Qualitative comparison on paired real-image relighting examples from the MIT Multi-Illumination dataset. Each row corresponds to a scene relit under a target illumination. Our method more closely matches the relit ground truth in global color tone, local shading, and cast-shadow appearance.
    }
    \label{fig:supp_setting5}
\end{figure*}

We show additional qualitative comparisons on synthetic and real-world relighting data in Figs.~\ref{fig:supp_setting1}-\ref{fig:supp_setting5}. \textbf{We also provide a project webpage containing video relighting comparisons at \url{http://zixinguo.me/lightcrafter/}}.

\subsection{Artifact-Matched Training Data}
\label{supp:artifact_matched_data}

\begin{figure*}[t]
    \centering
    \includegraphics[width=\textwidth,clip=true,trim=0 0mm 0mm 0mm]{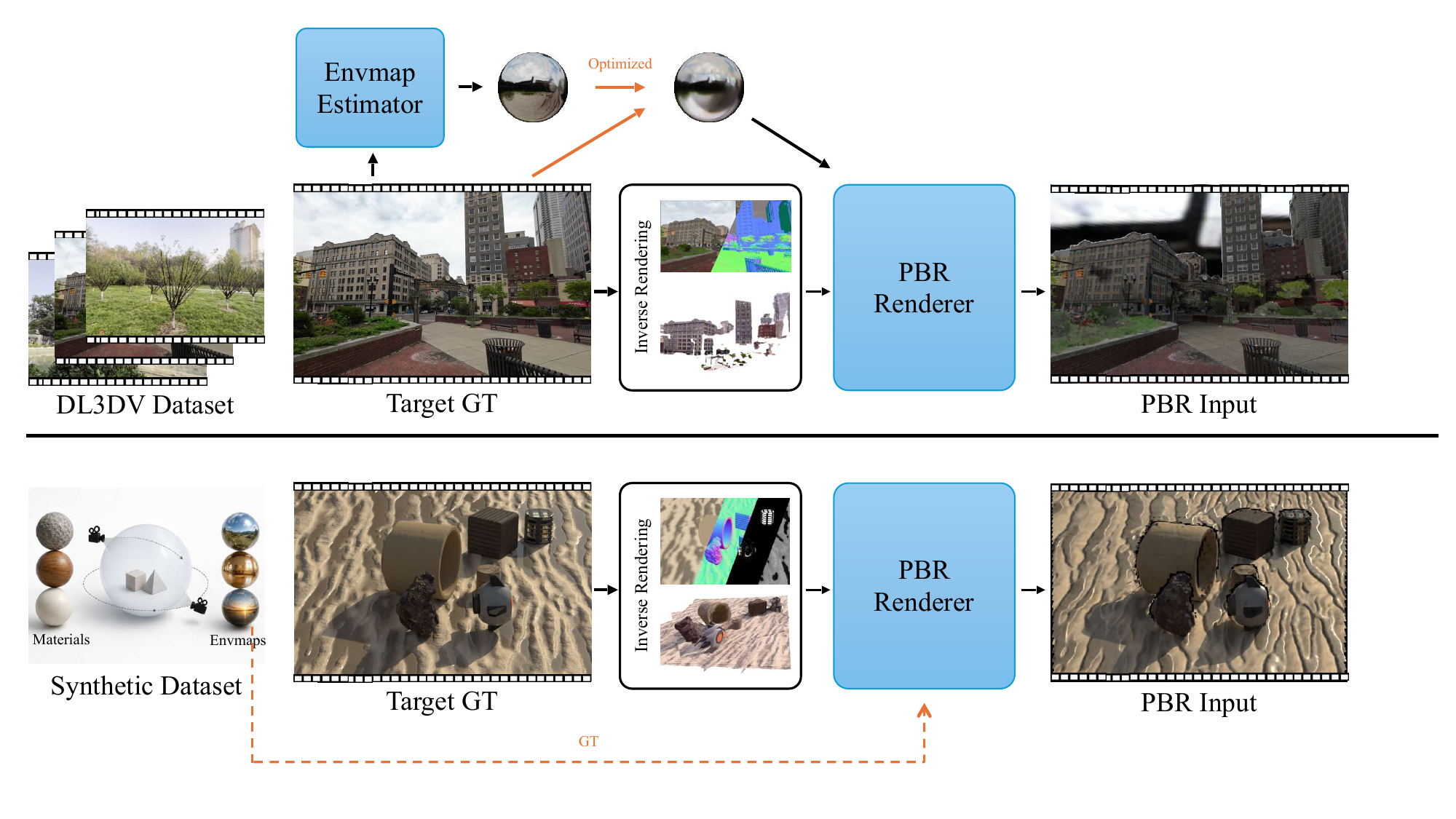}
    \squeeze{-5mm}
    \caption{\textbf{Synthetic and Real-World Training Data Curation.}
    }
    \label{fig:datacuration}
\end{figure*}

Our training data has \revision{three} complementary sources. The synthetic branch provides paired ground-truth relighting supervision, the \revision{DL3DV} branch exposes the model to in-the-wild reconstruction and rendering artifacts\revision{, and the MIT branch provides paired real relighting supervision}. Data curation pipeline diagram is provided in
Fig.~\ref{fig:datacuration}\revision{, and Table~\ref{tab:data_split} summarizes the training and evaluation splits}.

{\revstart
\begin{table}[h]
\centering
\small
\caption{\textbf{Training and evaluation splits.} All full-model results use the complete training set; ``w/o MIT'' removes the 750 MIT clips.}
\label{tab:data_split}
\begin{tabular}{@{}p{2.4cm}p{5.4cm}p{4.9cm}@{}}
\toprule
Source & Training & Evaluation \\
\midrule
Synthetic & 400 scenes / $\sim$3,200 clips; four HDR maps per scene, diverse camera and object motion; paired video relighting supervision. & 50 unseen scenes / 200 videos, each under four target HDR maps. \\
DL3DV~\citep{ling2024dl3dv} & 100 scenes / 1,200 clips; source-reconstruction pseudo-pairs rendered under the optimized source illumination and supervised by the original video. & 30 held-out scenes / 30 clips; relit with Light-A-Video, then relit back and compared with the original. \\
MIT~\citep{murmann19} & 30 scenes / 750 clips; captures under 25 flash directions, repeated into static 49-frame clips. & 10 scenes / 250 clips from the held-out ``everett'' residence. \\
\bottomrule
\end{tabular}
\end{table}

\paragraph{Train/evaluation differences.} Synthetic training and evaluation scenes are generated independently using disjoint Objaverse asset IDs and different camera and scene motion configurations. Their HDR maps are sampled from the same Poly Haven pool, following DiffusionRenderer~\citep{liang2025diffusionrenderer} and UniRelight~\citep{he2025unirelight}, so the test objects are novel but the test illumination distribution is shared, as for these baselines. DL3DV is split at the original-video level, so no evaluation scene contributes training clips; training uses source reconstruction, whereas evaluation uses the round-trip protocol of \citet{liu2025lightx}. MIT follows the official building-level split, with no room overlap between training and the held-out ``everett'' residence.
\par}

\paragraph{Synthetic Paired Videos.}

We curate and train on a dataset of paired videos and PBR renderings of size $49 \times 720 \times 480$,  constructed from estimated material buffers, camera poses, and depth rendered under target HDR environment maps. To construct the synthetic dataset we build compositional 3D scenes from Objaverse~\citep{objaverse} assets, primitives, and textured ground planes in Blender, and render videos under a target illumination randomly selected from a publicly available dataset~\citep{polyhaven, cctexturekey} using three camera motion configurations: a camera orbiting a static scene, a camera moving in a dynamic scene, and a static camera with rotating illumination. 

Specifically, each scene contains three filtered Objaverse~\citep{objaverse} objects and three primitive shapes placed on a textured ground plane under HDR environment illumination. Primitive objects and ground planes are assigned procedural or scanned materials sampled from texture libraries~\citep{polyhaven, cctexturekey}, while Objaverse assets provide diverse object geometry and appearance. This combination gives controllable ground-truth rendering while exposing the model to heterogeneous shapes, materials, shadows, and inter-object occlusions.

We generate 400 synthetic scenes. For each scene, we render multiple 49-frame camera trajectories under four lighting conditions. We define four elevation groups (upper, middle, lower, random), with the camera starting point sampled from one of four horizontal quadrants around the scene. Along each trajectory, the camera rotates horizontally by one degree per frame while distance to the scene and elevation are randomly varied. We additionally include a camera oscillation trajectory to model non-monotonic motion. This trajectory design exposes the refiner to view-dependent effects, irregular sampling patterns similar to handheld capture, and varying visibility, depth, and reconstruction errors.

For each synthetic configuration $q$ and target illumination $\mathbf{L}_\mathrm{target}$, we render a clean target video $\I_{\text{target}}(q,\mathbf{L}_\mathrm{target})$ using the Blender ground-truth scene. The corresponding proxy is \emph{not} rendered from ground-truth G-buffers; instead, we pass the source video through the same reconstruction stack $\Phi$ used at inference:
\begin{equation}
\widetilde{\mathcal{S}}^q = \Phi(\I_{\text{source}}^q),
\end{equation}
and render the target-light proxy from the recovered scene state. A direct use of $\widetilde{\mathcal{S}}$ can introduce a systematic mismatch with the Blender target because monocular reconstruction may have an arbitrary coordinate frame, camera scale, and depth scale; this would change the effective lighting direction and cast-shadow length. We therefore apply a low-dimensional alignment step:
\begin{equation}
\widehat{\mathcal{S}}^q = \operatorname{Align}\big(\widetilde{\mathcal{S}}^q,\, \mathcal{S}_{\text{gt}}^q\big),
\end{equation}
which estimates a global similarity transform and depth-scale correction from the synthetic ground-truth coordinate system. This alignment does \emph{not} replace the recovered buffers with ground truth. The local geometry errors, holes, noisy materials, inaccurate normals, and missing thin structures are preserved.

\paragraph{Real-World Video Pairs.}

To construct the training pairs from real-world videos, we use both the DL3DV~\citep{ling2024dl3dv} and the MIT Multi-Illumination~\citep{murmann19} datasets. From DL3DV, we sample 1{,}200 real-video chunks, each containing 49 frames. From the MIT dataset, we sample 30 scenes, each captured under 25 flash directions, with a chrome ball and a grey sphere in every image. MIT scenes are single images, so every pipeline stage collapses to its image-level variant and one mesh is shared across the 25 lights; each image is tiled into a static 49-frame video clip. Neither dataset provides ground-truth illumination, so we recover a pseudo source illumination instead: holding the geometry, cameras, and material buffers in $\mathcal{S}$ fixed, we optimize $\mathbf{L}_{\text{source}}$ with Eq.~\eqref{eq:envmap_opt} of the main paper. The total variation loss $\mathcal{L}_\mathrm{TV}$ in Eq.~\eqref{eq:envmap_opt} penalizes per-pixel discontinuities in the latitude-longitude environment map. For an environment map $\mathbf{L} \in \mathbb{R}^{H_e \times W_e}$ with longitude index $u$ and latitude index $v$:
\begin{equation}
\mathcal{L}_\mathrm{TV}(\mathbf{L}) = \sum_{u,v} \big| \mathbf{L}_{u+1,v} - \mathbf{L}_{u,v} \big| + \big| \mathbf{L}_{u,v+1} - \mathbf{L}_{u,v} \big|,
\label{eq:tv}
\end{equation}
with horizontal wraparound at the longitudinal boundary ($u = W_e$ wraps to $u = 0$) to respect the spherical topology of the environment map. We apply this regularizer to $\log \mathbf{L}_{\text{source}}$ rather than $\mathbf{L}_{\text{source}}$ directly so that the penalty is scale-invariant across the dynamic range of HDR illumination, treating relative differences between bright sun pixels and dim sky pixels uniformly.

We initialize $\mathbf{L}_{\text{source}}$ from DiffusionLight~\citep{phongthawee2024diffusionlight} on the first frame, which we found accelerates convergence and avoids artifacts in the rendering. The illumination is parameterized as a $64 \times 128$ latitude-longitude HDR environment map in log space and exponentiated at the renderer call site. We additionally optimize three scalar parameters (global yaw, pitch, and exposure) that absorb gauge ambiguities in the recovered geometry. We use Adam (learning rate $2 \times 10^{-3}$, default betas) for 2{,}000 iterations per clip. At each iteration, we sample one frame from every fourth frame to bound memory, and render $\mathcal{R}(\mathcal{S}^i, \mathbf{L}_{\text{source}})$ at $480 \times 720$ with 2{,}048 light samples and mesh-traced shadow visibility. The rendered output is Reinhard-tonemapped before the $\ell_1$ residual to keep the loss in display range. We set $\lambda_{\text{TV}} = 10^{-3}$, large enough to suppress pixel-level speckles while still allowing low-frequency directional structure (sun position, sky-vs-floor gradient) to emerge from the data. Optimization runs independently per scene on a single A6000 GPU. Once converged, the optimized environment map is held fixed, and the renderer produces all 49 frames of the pseudo-input video $\hat{\I} = \mathcal{R}(\mathcal{S}, \mathbf{\hat{L}}_{\text{source}})$ paired with the original $\I$ for training.

{\revstart
\paragraph{Role and limits of source-light estimation.}
\label{supp:source_light}
Estimating an HDR environment map from a single LDR image is inherently ambiguous. In our pipeline, however, the DiffusionLight prediction serves only as an initialization: because the estimated G-buffers are consistent across lighting conditions, the multi-frame differentiable rendering above refines one shared environment map to reproduce the observed frames. We empirically found that it substantially reduces sensitivity to the initial estimate, and it is used only to build real-world training pairs. At inference, source-light estimation is not required, since the estimated G-buffers are rendered directly under the user-specified target illumination, so errors in the source-light estimate do not propagate to the requested lighting, and the remaining inference errors come from the estimated geometry and materials analyzed in Sec.~\ref{sec:error_source}.
\par}

\subsection{Video Diffusion Proxy Refiner and Long-Video Inference}
\label{supp:refiner}

\paragraph{Robust Conditioning via Input Noise.}
To prevent the refiner from over-fitting to artifact patterns specific to a
particular IR run, the encoded PBR proxy is corrupted with per-batch Gaussian
noise before encoding:
\begin{equation}
  \tilde{\mathbf{P}} \;=\; \mathbf{P} + \sigma \cdot \boldsymbol{\eta}, \qquad
  \boldsymbol{\eta} \sim \mathcal{N}(\mathbf{0}, \mathbf{I}), \qquad
  \log \sigma \sim \mathcal{N}(-3,\; 0.5),
\end{equation}
sampled independently per training step. The log-normal schedule covers
$\sigma$ from $\sim$0.005 to $\sim$0.13 with the bulk near 0.05.

\paragraph{Inference.}
For sequences within the training window we run the full DPM solver for
50 denoising steps with classifier-free guidance scale $\omega = 6.0$ and
time-step spacing \texttt{trailing}. The empty prompt is reused as both the
unconditional and conditional cross-attention key, so guidance acts solely
on the input latent. \revision{Runtime is reported in Appendix~\ref{supp:runtime}.}

\paragraph{Inference on Long Videos via Temporal Tiling.}
\label{supp:long_video}

\paragraph{Long-video overlap fusion.}
For videos longer than the training window, we use overlap-fused temporal tiling as the default long-video inference procedure. 
This fusion is performed inside each diffusion denoising step, rather than as a post-processing step on independently decoded clips. 
Let $N$ be the number of pixel frames, and let
$\mathbf{z}_t \in \mathbb{R}^{B\times C\times N'\times H'\times W'}$
denote the full-sequence noisy latent at denoising timestep $t$, where $B$ is the batch size, $C$ is the latent channel dimension, and $N'$, $H'$, and $W'$ are the temporal and spatial latent resolutions after VAE encoding. 
Let $\mathbf{z}^{\mathbf{P}}\in \mathbb{R}^{B\times C\times N'\times H'\times W'}$ be the latent encoding of the corresponding full PBR proxy video.

Our model is trained on clips of $T=49$ pixel frames. 
For long videos, we use overlapping temporal windows whose decoded support corresponds to $T=49$ pixel frames, with a pixel-space overlap of $\kappa=12$ frames. 
We denote the corresponding latent window length and latent overlap by $T'$ and $\kappa'$, respectively, and define the latent stride as $\Delta'=T'-\kappa'$. 
The temporal windows are
\begin{equation}
W_j = [s_j,\, s_j+T'), 
\qquad
s_j = j\Delta',
\qquad
j=0,\ldots,K-1,
\end{equation}
with the final window adjusted, when necessary, to ensure coverage of the full latent sequence up to index $N'$. 
Let $C_j$ be the crop operator that extracts window $W_j$ from a full latent sequence, and let $C_j^\top$ be the corresponding scatter operator that writes a window prediction back into the full latent sequence.

At each denoising timestep, we apply the trained video refiner to every temporal window:
\begin{equation}
\hat{\boldsymbol{\epsilon}}_{\theta,j}
=
\epsilon_{\theta}
\!\left(
\left[
C_j\mathbf{z}_t
\,\Vert\,
C_j\mathbf{z}^{\mathbf{P}}
\right]_{\mathrm{ch}},
t
\right),
\label{eq:tiled_window_prediction}
\end{equation}
where $[\cdot\Vert\cdot]_{\mathrm{ch}}$ denotes channel-wise concatenation and
$\hat{\boldsymbol{\epsilon}}_{\theta,j}\in
\mathbb{R}^{B\times C\times T'\times H'\times W'}$
is the noise prediction for window $W_j$. 
We then fuse the per-window predictions in noise-prediction space:
\begin{equation}
\hat{\boldsymbol{\epsilon}}_{\theta}
=
\frac{
\sum_{j=0}^{K-1}
C_j^\top
\!\left(
\mathbf{w}_j \odot \hat{\boldsymbol{\epsilon}}_{\theta,j}
\right)
}{
\sum_{j=0}^{K-1}
C_j^\top(\mathbf{w}_j) + \delta
},
\label{eq:tiled_fusion}
\end{equation}
where $\mathbf{w}_j\in\mathbb{R}^{1\times 1\times T'\times 1\times 1}$ is a temporal tent weight broadcast over batch, channel, and spatial dimensions, and $\delta$ is a small constant for numerical stability. 
Finally, we apply a single global scheduler update to the full latent sequence:
\begin{equation}
\mathbf{z}_{t-1}
=
\mathrm{Step}
\!\left(
\mathbf{z}_t,
\hat{\boldsymbol{\epsilon}}_{\theta},
t
\right).
\label{eq:tiled_scheduler_step}
\end{equation}
Because all windows contribute to one fused noise prediction before the scheduler update, the method maintains a single coherent latent trajectory over the full video, rather than denoising and stitching independent clips.

{\revstart
\subsection{Runtime and Memory}
\label{supp:runtime}
Table~\ref{tab:runtime} gives runtime and peak memory per component on a single NVIDIA A6000 (48\,GB) for a 49-frame clip at 480$\times$720. Inference does not need light estimation and per-video lighting optimization, since the target illumination is user-specified.

\begin{table}[h]
\centering
\small
\caption{\textbf{Per-component cost.} $^*$ = mutually independent, run concurrently; t = training-pair construction only; i = inference only.}
\label{tab:runtime}
\begin{tabular}{@{}lcc@{}}
\toprule
Component & Runtime & Peak VRAM \\
\midrule
Ground-truth target$^{*}$ (t) & 1m 00s & -- \\
Appearance reconstruction~\citep{liang2025diffusionrenderer}$^{*}$ & 8m 47s & 25.8 GB \\
Geometry reconstruction~\citep{li2024megasamaccuratefastrobust}$^{*}$ & 5m 17s & 17.9 GB \\
Per-frame mesh$^{*}$ & 1m 12s & -- \\
Light estimation~\citep{phongthawee2024diffusionlight}$^{*}$ (t) & $\sim$1m 50s & 17.4 GB \\
Lighting optimization~\citep{jakob2022mitsuba3} (t) & $\sim$7m 00s & 14.4 GB \\
Relighting via PBR~\citep{jakob2022mitsuba3} & 0m 58s & -- \\
Diffusion refinement~\citep{yang2025cogvideoxtexttovideodiffusionmodels} (i) & 5m 33s & 21.1 GB \\
\bottomrule
\end{tabular}
\end{table}

Taking the maximum over concurrent branches plus the sequential steps that follow, building one training pair costs 9m 45s for synthetic data and 16m 45s for DL3DV. MIT scenes are single images tiled into static clips, so every stage collapses to its image-level variant and one mesh is shared across the 25 lights, giving about 7m 35s per MIT pair. At inference, appearance and geometry reconstruction run in parallel, followed by relighting and refinement, so one 49-frame 480$\times$720 video costs $\max(8\text{m}\,47\text{s},\, 6\text{m}\,29\text{s}) + 0\text{m}\,58\text{s} + 5\text{m}\,33\text{s} = 15\text{m}\,18\text{s}$ end-to-end on a single A6000.
\par}

\subsection{Baseline Implementation}
\label{supp:baselines}
\revision{\paragraph{Evaluation settings.} All methods receive the same source inputs and target HDR environment map and are evaluated at 720$\times$480. For the 200-frame synthetic evaluation sequences, DiffusionRenderer and UniRelight were trained on 57-frame chunks and Light-X on 49-frame chunks; since each model produces its best quality at its training length, we run every method at its own training length and concatenate the outputs before computing metrics over the same complete sequence. For DiffusionRenderer and UniRelight, we rotate the target environment map according to the recovered camera trajectory to keep illumination fixed in world coordinates. All diffusion baselines use 50 denoising steps, and all methods are scored on raw outputs without scale, exposure, or color alignment.}

\begin{figure*}[t]
    \centering
    \includegraphics[width=\textwidth,clip=true,trim=0 0mm 0mm 0mm]{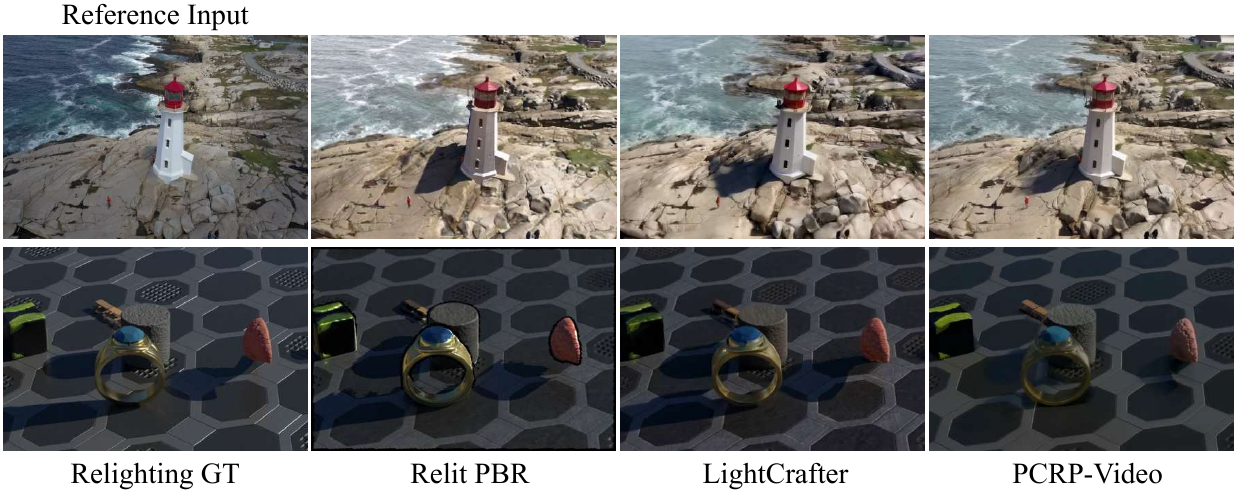}
    \vspace*{-5mm}
    \caption{Qualitative comparison with the PCRP-video extension baseline on static-scene orbiting-camera relighting. Both scenes are relatively static, with the camera orbiting around the subject. 
LightCrafter better captures shadow appearance, produces higher-quality relit videos, and more closely matches the ground truth in the synthetic setting.
    }
    \label{fig:oursvspcrp}
\end{figure*}

\paragraph{PCRP-Video.}
PCRP \citep{careagaRelighting} is originally designed for image relighting, and no public implementation is available at the time of writing. 
We therefore implement a video extension for evaluation by following its high-level pipeline: intrinsic decomposition and geometry recovery, PBR rendering under target illumination, and neural refinement. 
For the video setting, we replace the image-level decomposition and monocular geometry modules with our video inverse-rendering pipeline and MegaSAM-based geometry estimates. 
We train this PCRP-style video baseline on the same synthetic and real-world data as our model, except we use their self-supervised curation approach, i.e., without ground-truth lighting. This baseline generally captures coarse relighting and follows the shading direction of the PBR proxy, but it is less effective at correcting structured illumination artifacts. 
In particular, it often fails to recover accurate cast shadows and high-frequency light effects such as highlights, glare, and specular reflections, as indicated in 
Fig.~\ref{fig:oursvspcrp}

{\revstart
\subsection{Additional Experiments and Analyses}
\label{supp:additional}

\paragraph{Per-dataset real-world results.} MIT Multi-Illumination has paired relit captures of the same scene, so its metrics measure relighting accuracy. DL3DV has no relit ground truth, so its round-trip metrics measure preservation and temporal stability rather than relighting fidelity; because Light-A-Video already loses fine detail, much of the residual error is inherited upstream, and even the strongest methods saturate near 19--20\,dB. Only DL3DV is video, so temporal metrics on real data would be DL3DV-only. Averaging the two datasets in Table~\ref{tab:main_relighting} recovers the combined real-world numbers of Table~\ref{tab:ablation_relighting}. For the qualitative real-world examples, we deliberately choose high-contrast HDR maps with strong directional light, which make fine cast shadows from foliage and irregular geometry easy to judge; the stronger shading is already present in the PBR rendering and follows from the source--target illumination pair rather than from a sim-to-real gap.

\paragraph{Robustness to inverse-rendering errors.}\label{sec:error_source} Inverse rendering de-lights the source video, so it passes illumination-invariant quantities (basecolor, normal, roughness, metalness) downstream, and the illumination used for relighting is the user-specified target rather than an estimate. Errors therefore enter the PBR rendering through geometry and material. Therefore, for an analysis of our refiner's robustness to upstream errors, we isolate synthetic scenes where ground-truth buffers exist: we fix one to ground truth, estimate the other, and score both the PBR rendering and the refined output against paired target-light renders (Table~\ref{tab:error_source}). Geometry error damages the PBR rendering most (19.86 vs.\ 22.46\,dB), yet is also the most correctable: the refiner recovers +4.79\,dB (LPIPS $-$0.077) there, versus +1.43\,dB (LPIPS $-$0.018) under material error, where the PBR rendering is already strong and refinement mainly improves perceptual quality. Missing boundaries, thin structures, visibility mistakes, and broken shadows are spatially localized artifacts, which learned refinement repairs best. Both settings converge to similar quality after refinement (LPIPS $\approx$ 0.21--0.23). The full pipeline refines slightly better than the ground-truth-mesh setting, which we hypothesize is because the refiner is trained on proxies built from predicted meshes.

\begin{table}[h]
\centering
\small
\caption{\textbf{Isolating inverse-rendering error sources} on synthetic scenes. Each row fixes one of the two inverse-rendering outputs (mesh, G-buffer) to ground truth and estimates the other. For every metric we report the PBR rendering before refinement, the refined output, and the change $\Delta$ due to refinement.}
\label{tab:error_source}
\setlength{\tabcolsep}{2.2pt}
\footnotesize
\begin{tabular}{@{}lll ccc ccc ccc@{}}
\toprule
& & & \multicolumn{3}{c}{PSNR $\uparrow$} & \multicolumn{3}{c}{SSIM $\uparrow$} & \multicolumn{3}{c}{LPIPS $\downarrow$} \\
\cmidrule(lr){4-6} \cmidrule(lr){7-9} \cmidrule(lr){10-12}
Error source & Mesh & G-buffer & PBR & Refined & $\Delta$ & PBR & Refined & $\Delta$ & PBR & Refined & $\Delta$ \\
\midrule
Material only & GT & pred. & 22.46 & 23.89 & $+$1.43 & 0.7163 & 0.7165 & $+$0.0002 & 0.2526 & 0.2342 & $-$0.0184 \\
Geometry only & pred. & GT & 19.86 & 24.65 & $+$4.79 & 0.7127 & 0.7213 & $+$0.0086 & 0.2856 & 0.2091 & $-$0.0765 \\
Full pipeline & pred. & pred. & 19.59 & 24.16 & $+$4.57 & 0.7111 & 0.7366 & $+$0.0255 & 0.3106 & 0.2096 & $-$0.1010 \\
\bottomrule
\end{tabular}
\end{table}

\par}

{\revstart
\subsection{User Study Protocol}
\label{supp:userstudy}
We showed to the users 11 comparison sets.
Each of the 11 comparison sets pairs the source video and target illumination with five relit results in randomized order (Light-X, UniRelight, DiffusionRenderer, PCRP-Video, and ours) and asks for the single best result under four criteria: temporal consistency, relighting accuracy, overall visual quality, and scene-content preservation. This is a forced top-1 choice among five, so chance is 20\%. The videos are deliberately weighted toward out-of-distribution footage: 8 in-the-wild videos (Pexels stock footage and Sora 2 / Kling generations), 2 held-out synthetic scenes (one with a moving camera and one with dynamic lighting, both static scenes), and 1 held-out DL3DV scene. Sixteen participants answered all 44 questions, giving 704 judgments.

 Over 704 forced top-1 choices among five methods (chance 20\%), \OurMethod{} is the most preferred method on all four criteria (Table~\ref{tab:user_study}), most strongly on relighting accuracy (67.6\%, over three times chance); all rates are far above chance ($p < 10^{-5}$, one-sided binomial test).

\begin{table}[h]
\centering
\small
\caption{\textbf{User study.} Top-1 preference rate (\%) over 704 forced choices among five methods; chance is 20\%.}
\label{tab:user_study}
\begin{tabular}{@{}lcc@{}}
\toprule
Criterion & Ours & Each baseline (avg.) \\
\midrule
Temporal consistency & \textbf{60.8} & 9.8 \\
Relighting accuracy & \textbf{67.6} & 8.1 \\
Overall visual quality & \textbf{57.4} & 10.7 \\
Scene-content preservation & \textbf{45.5} & 13.6 \\
\midrule
All criteria & \textbf{57.8} & 10.5 \\
\bottomrule
\end{tabular}
\end{table}

\subsection{Indoor Light-Editing}
\label{supp:indoor}
In addition to outdoor lighting controlled by an environment map, we apply our method to indoor light-editing tasks outside our training-data distribution and compare it with indoor scene-specific baselines GR3EN~\citep{xing2026gr3en} and LuxRemix~\citep{liang2026luxremix}. 

In our pipeline, we place the edited lights at their 3D positions either by unprojecting user-specified 2D keypoints into the reconstructed scene (so the edited light must be visible in the video) or by placing them directly in Blender. GR3EN runs with its official weights, and its palette mask is matched to the ground-truth lamps. The public LuxRemix release contains only the SEVA~\citep{zhou2025stable} multi-view propagation weights, not its single-view editing stage. We therefore construct the reference view ourselves: the input middle frame with the light masks composited onto the fixtures as a stand-in edit, which LuxRemix then propagates to the remaining views.

We run both baselines against our method on in-the-wild footage where no ground truth exists. In the meeting room of Fig.~\ref{fig:indoor_meetingroom}, two neutral ceiling panels are recolored to blue and yellow; all methods receive the same fixture mask. Ours keeps each color on its own panel and casts blue and warm spills onto the corresponding walls while leaving the rest of the room intact. GR3EN instead tints the whole room warm and drops the blue light, and LuxRemix brightens the room roughly uniformly with almost no colored light reaching the scene. The edit is rendered into our PBR proxy before refinement, so the requested light color, position, and falloff are enforced physically rather than inferred from a mask, which is what gives us the finer light controllability visible on the project page.

\begin{figure}[h]
\centering
\includegraphics[width=\linewidth]{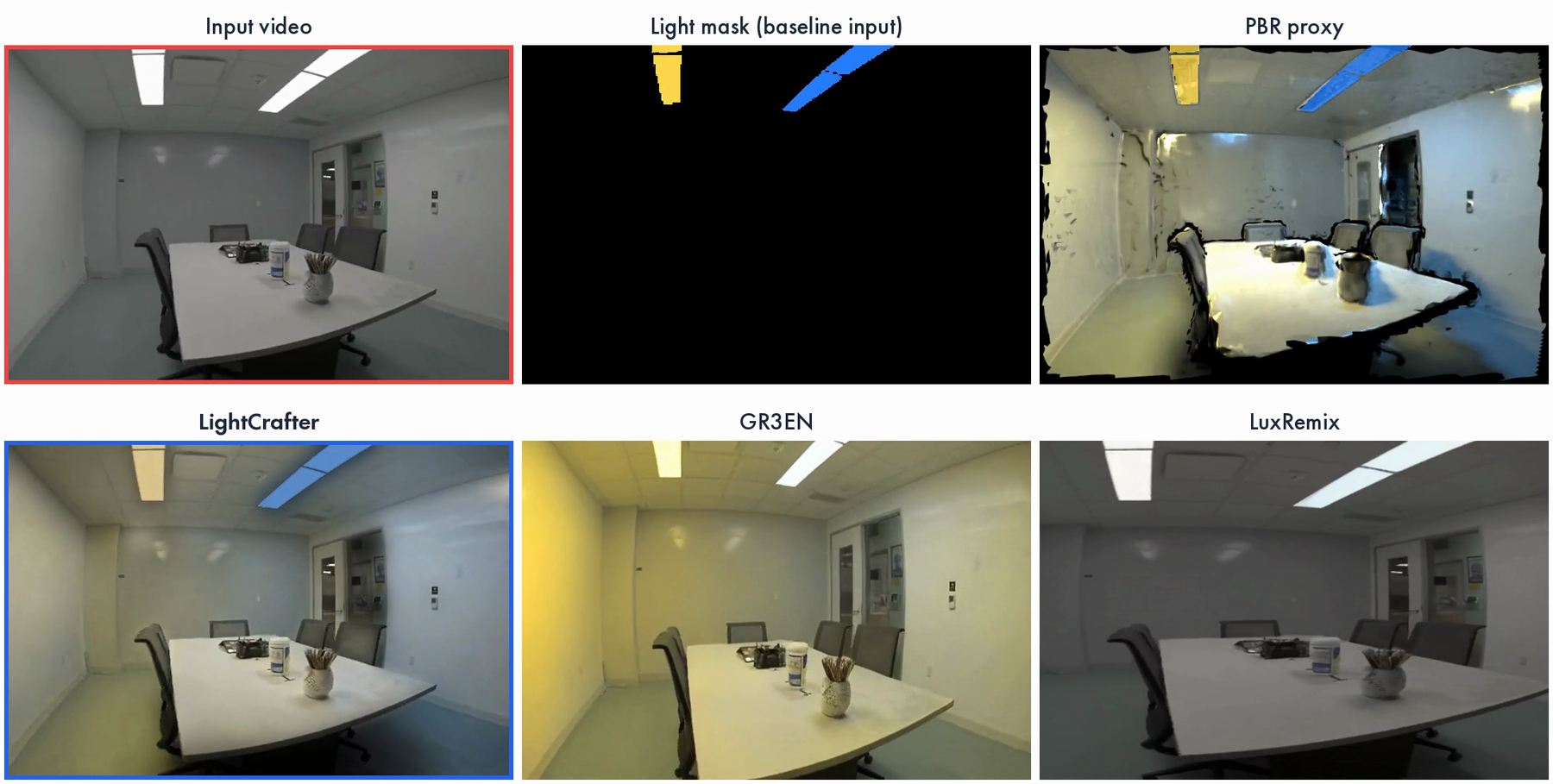}
\caption{\textbf{In-the-wild light editing (meeting room, no ground truth).} Two neutral ceiling panels are recolored to blue and yellow; the same fixture mask is given to every method. Ours keeps each color on its own panel with blue and warm spill on the corresponding walls; GR3EN tints the whole room warm and drops the blue light; LuxRemix brightens the room uniformly with little colored light. Full videos are on the project page.}
\label{fig:indoor_meetingroom}
\end{figure}

\paragraph{Comparison of the control interfaces.} Both baselines select a light by drawing a 2D mask over its fixture, so a light in the next room or behind the camera cannot be selected, and remixing can only rebalance lights that already illuminate the capture; neither can add a new source, change the environment lighting, or combine the two as in the car-headlight insertion of Fig.~\ref{fig:applications}. The same restriction means these indoor light-editing interfaces do not transfer to outdoor envmap relighting, which limits their control formulation rather than our evaluation. Our interface operates in 3D: lights are explicit entities in the PBR proxy with position, orientation, intensity, color, and emission profile, so a source is selected by identity, an out-of-view light is handled like a visible one, and because the edit is rendered rather than remixed, the same interface covers inserting sources, swapping the environment map, and dynamic target lighting, with shadows and global illumination following from the edit.


\subsection{Failure Cases and Additional Analyses}
\label{supp:failures}
\revision{\paragraph{Limitations and failure cases.} Despite training on artifact-matched proxies, the video refiner of LightCrafter can fail when severe inverse-rendering errors obscure essential scene information. Under extreme camera or object motion and strong motion blur, per-frame geometry becomes unreliable, and the refiner can hallucinate plausible but incorrect structure; scenes with very weak parallax or fine thin structures produce proxies with gross depth errors, missing geometry, or large visibility holes that cannot be filled from the proxy alone; and highly specular, transparent, or metallic surfaces remain the hardest materials to recover. On very long sequences, small per-window inconsistencies can also accumulate across chunks into a slow appearance drift. Stronger geometric priors or multi-view cues, richer PBR representations, and jointly optimizing inverse rendering with refinement are natural directions for pushing these boundaries further.}

\par}

\end{document}